\documentclass[sigconf,balance=false]{acmart}
\usepackage{xspace}
\usepackage{algpseudocode}
\usepackage{algorithm}
\usepackage{booktabs}
\usepackage{graphicx}
\usepackage{multirow}
\usepackage{xcolor}
\AtBeginDocument{%
  }

\setcopyright{acmlicensed}
\copyrightyear{2018}
\acmYear{2018}
\acmDOI{XXXXXXX.XXXXXXX}

\acmConference[Preprint]{Make sure to enter the correct
  conference title from your rights confirmation email}{}{}

\acmISBN{978-1-4503-XXXX-X/2018/06}

\setcopyright{none}
\renewcommand\footnotetextcopyrightpermission[1]{}

\newcommand{\icon}{\raisebox{-2pt}{\includegraphics[width=1.47em]{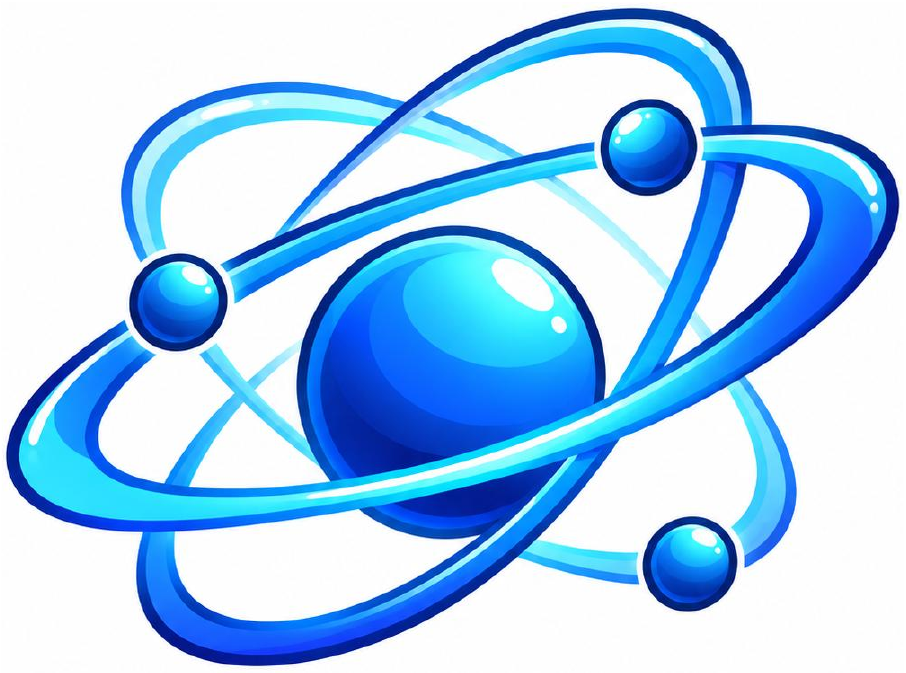}}\xspace}

\begin{document}


\title[\icon ORBIT: Autoregressive Time Series Generation via Next Frequency-Band Prediction]{
\icon ORBIT: Autoregressive Time Series Generation via \\
Next Frequency-Band Prediction}

\author{Chunyi Hou}

\affiliation{%
  \institution{School of Data Science and Engineering, East China Normal University}
  \city{Shanghai}
  \country{China}
}

\author{Xiangfei Qiu}
\affiliation{%
  \institution{School of Data Science and Engineering, East China Normal University}
  \city{Shanghai}
  \country{China}}
\author{Hanyin Cheng}
\affiliation{%
  \institution{School of Data Science and Engineering, East China Normal University}
  \city{Shanghai}
  \country{China}}


\author{Chenchen Sun}
\affiliation{%
  \institution{Chang'an University}
  \city{Xi'an}
  \country{China}
}

\author{Yisheng An}
\affiliation{%
  \institution{Chang'an University}
  \city{Xi'an}
  \country{China}
}

\author{Yang Shu}
\affiliation{%
 \institution{School of Data Science and Engineering, East China Normal University}
 \city{Shanghai}
 \country{China}}

\author{Jilin Hu}
\affiliation{%
  \institution{School of Data Science and Engineering, East China Normal University}
  \city{Shanghai}
  \country{China}}

\author{Bin Yang}
\affiliation{%
  \institution{School of Data Science and Engineering, East China Normal University}
  \city{Shanghai}
  \country{China}}


\begin{abstract}
Deep generative models provide an effective modern solution to alleviating the problem of data scarcity by synthesizing realistic time series.
However, despite the considerable progress achieved by existing deep time series generators, two key challenges remain:
(1) learning hierarchical representations for heterogeneous
temporal patterns;
and (2) coordinating the generation process of heterogeneous temporal dynamics.
To address these key bottlenecks, in this work, we propose \textbf{ORBIT}, a unified deep generative framework for time series.
Specifically, we develop two key components: the Frequency-Hierarchical VQ-VAE (FH-VQ-VAE) and the Frequency-Band Autoregressive Transformer (FBAR).
The former learns hierarchical representations by decomposing latent temporal representations into different frequency bands and applying residual quantization with increasing token budgets.
The latter establishes a novel low-to-high frequency autoregressive direction, yielding a time series generation process that first constructs global temporal structures and then incrementally incorporates fine-grained temporal details.
These compact designs not only disentangle heterogeneous temporal patterns for more effective representation learning but also mitigate potential interference among different temporal dynamics during generation.
Extensive experiments on multiple challenging real-world benchmarks demonstrate that ORBIT achieves state-of-the-art overall performance while exhibiting superior inference efficiency compared with advanced baselines.

\end{abstract}

\ccsdesc[500]{Computing methodologies~Artificial intelligence}

\keywords{Time Series Generation}


\maketitle

\section{Introduction}
The rapid advancement of complex intelligent systems has led to a substantial growth in the volume of time series data, which are now ubiquitous across many critical real-world domains, including meteorology~\citep{weatherforecasting1}, transportation~\citep{trafficsurvey}, healthcare~\cite{titd}, energy~\citep{energysurvey}, and AIOps~\citep{aiops}.
In modern networking, mobile communications, and Web of
Things (WoT) ecosystems, such data capture evolving traffic
patterns, service demands, and sensor measurements.
This provides a strong foundation for critical downstream tasks, including informed decision-making and anomaly detection~\cite{tfb,tab}.
Unfortunately, access to these valuable data sources is often constrained by external factors, such as privacy concerns and extreme events~\citep{zhang2026survey}.
Deep generative models offer a promising solution to this problem by generating realistic synthetic time series~\citep{tsgbench}.
\begin{figure}
    \centering
    \includegraphics[width=1.\linewidth]{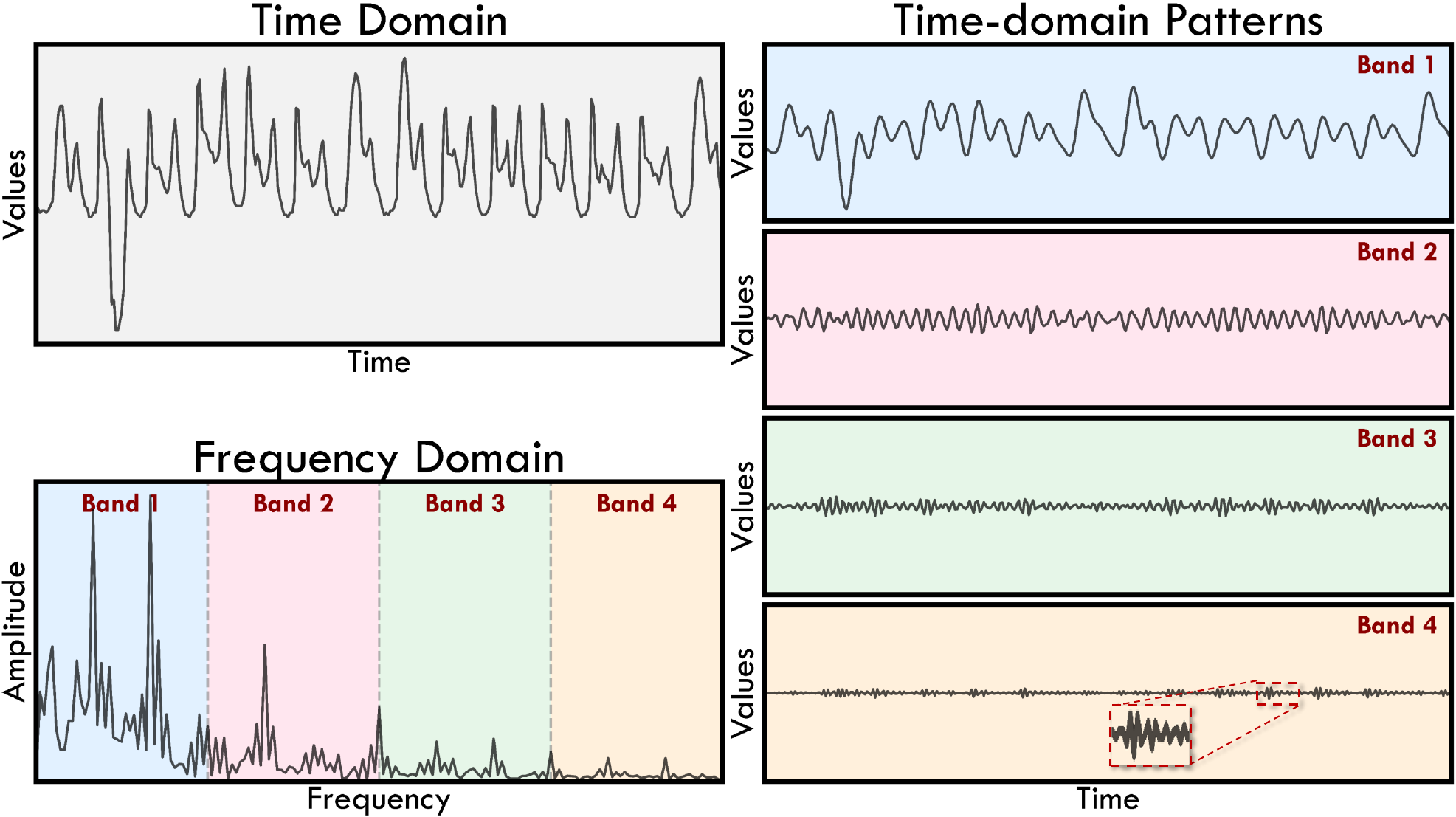}
    \caption{Different frequency bands correspond to distinct temporal dynamics, with low-frequency components capturing smooth and slow variations, while high-frequency components characterize rapid temporal fluctuations.}
    \label{fig:1}
\end{figure}

\begin{figure}[h]
    \centering
    \includegraphics[width=0.83\linewidth]{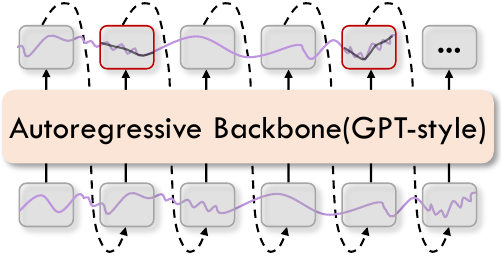}
    \caption{ Time Series Generation with a Standard Autoregressive Backbone.
    At certain inference steps (highlighted by the \textcolor{red}{red boxes}), the model is needed to jointly generate coarse-grained global trends and fine-grained local fluctuations.}
    \label{fig:2}
\end{figure}
In general, developing a deep generative model for time series involves two main stages: 
(1) learning a tokenizer that can transform original time series to compact and expressive temporal representations; 
and (2) training a generative model to fit the distribution of these tokens in the latent space.
This \emph{two-stage modeling paradigm} has gradually emerged as a prevalent and mainstream approach for deep time series generation~\citep{sdformer,msdformer,polarformer,sdflow}.

However, real-world time series exhibit heterogeneous temporal patterns that are superimposed in the observation space. 
Existing deep time generators typically entangle these patterns within a single representation and fail to explicitly account for their distinct temporal dynamics during autoregressive generation.
For these reasons, developing a method that can effectively disentangle heterogeneous temporal patterns and coordinate their generation is crucial but challenging.
More specifically, achieving these objectives requires overcoming the following two  major obstacles:

\textbf{Obstacle 1: Difficulty in Learning Hierarchical Representations within Time Series.}
Due to the heterogeneity of temporal patterns, time series often contain multiple dynamic patterns evolving at different levels~\citep{duet,timemosaic,graphpatchformer}.
These heterogeneous patterns are superimposed in the numerical observation space, making their underlying dynamics difficult to disentangle and model separately.
As shown in Figure~\ref{fig:1}, Fourier analysis provides an intuitive means from a frequency-domain perspective to deeply understand them: 
low-frequency components mainly contain slow and smooth variations, whereas high-frequency components characterize rapid oscillations and fine-grained changes.
Without explicitly modeling these differences in temporal dynamics, integrating diverse patterns into a single representation may make it difficult to simultaneously capture global structures and local temporal details.
Therefore, how to explicitly distinguish and learn different temporal patterns to construct hierarchical representations that preserve both global structures and local details remains a key obstacle.

\textbf{Obstacle 2: Difficulty in Coordinating the Generation of Heterogeneous Temporal Dynamics.}
Conventional (GPT-style) autoregressive paradigms typically produce time series sequentially along the temporal dimension, either point by point or segment by segment~\citep{sdformer}. 
Their autoregressive direction strictly follows the left-to-right ordering, analogous to natural language modeling, and is primarily determined by the chronological order of observations~\citep{ntp}.
Although this approach exhibits reasonable generative capability, each inference step is still required to jointly generate content containing both slowly varying structures and rapidly fluctuating changes, which potentially causes interference between these heterogeneous temporal patterns, as illustrated in Figure~\ref{fig:2}.
Accordingly, another key obstacle is how to coordinate the generation of slowly varying global structures and rapidly fluctuating temporal details while preserving their dependencies and the coherence of the resulting time series.

In this paper, we address the above two key bottlenecks by introducing a unified deep generative framework, named \textbf{ORBIT}, which performs aut\textbf{O}reg\textbf{R}essive time series generation via next frequency-\textbf{B}and pred\textbf{I}c\textbf{T}ion.
Unlike existing methods, ORBIT explicitly disentangles components across different frequency-bands to construct hierarchical representations, and further designs a novel autoregressive direction that progresses from low- to high-frequency bands.

Specifically, \textbf{to address the first obstacle}, we propose a Frequency-Hierarchical VQ-VAE (FH-VQ-VAE).
FH-VQ-VAE decomposes latent temporal representations into distinct frequency bands and tokenizes these representations in a residual manner from low- to high-frequency bands with increasing token budgets, constructing hierarchical discrete latent representations.
In doing so, by progressively incorporating higher-frequency components
while refining earlier quantization errors, this design facilitates the disentangled learning of heterogeneous temporal patterns within time series.

On the other hand, \textbf{to address the second obstacle}, we design a Frequency-Band Autoregressive Transformer (FBAR), which \emph{reformulates autoregressive time series generation as a next frequency-band prediction task}.
FBAR generates frequency bands autoregressively from low to high frequencies, first establishing global temporal structures and then incrementally incorporating finer temporal details.
By conditioning each subsequent band on previously generated bands, FBAR explicitly models cross-band dependencies, enabling the coordinated generation of heterogeneous temporal dynamics.

Experimentally, we conduct extensive experiments to comprehensively evaluate ORBIT in both in-domain and cross-domain generation settings.
The results demonstrate that ORBIT consistently outperforms advanced baselines in generation performance, while offering substantial advantages in inference efficiency.

Overall, our main contributions are summarized as follows:

\begin{itemize}
    \item 
    We introduce ORBIT, a unified deep generative framework that enables high-quality and high-fidelity time series generation by combining hierarchical representation learning with a novel next frequency-band prediction paradigm.

    \item 
    We design FH-VQ-VAE, which combines explicit frequency-band
decomposition with hierarchical residual quantization to learn
discrete representations of heterogeneous temporal patterns,
capturing both global temporal structures and local
fine-grained dynamics.

    \item 

    We propose FBAR, which reformulates autoregressive time series generation as a next frequency-band prediction task, establishes a low-to-high-frequency autoregressive direction, and models cross-band dependencies to coordinate the generation of heterogeneous temporal dynamics.

    \item
    Extensive experiments are conducted on multiple challenging real-world benchmarks, whose results demonstrate that ORBIT achieves state-of-the-art overall generation performance while offering superior inference efficiency.

\end{itemize}







\section{Related Works}

\subsection{Deep Time Series Generation}
Early works of deep time series generation primarily focus on Generative Adversarial Networks (GANs), such as TimeGAN~\citep{TIMEGAN} and GT-GAN~\citep{gtgan}.
Contemporaneous studies also explore VAE-based approaches~\citep{vae}, such as TimeVAE~\citep{timevae}.
Yet, GANs and VAEs inherently suffer from issues such as training instability and posterior collapse, respectively, making such neural architecture-driven models difficult to train and therefore limiting their ability to generate high-quality synthetic time series.
With continued research progress, two-stage deep generative models are emerging as a more advanced and increasingly prevalent paradigm~\citep{zhang2026survey}.
Representative efforts such as SDformer~\citep{sdformer} first introduce the two-stage deep generative modeling paradigm to time series generation community, where a VQ-VAE~\citep{vqvae} learns discrete representations and a GPT-style autoregressive backbone~\citep{gpt4} subsequently models the distribution over the resulting discrete tokens.
Recently, several methodologies extend this paradigm to multi-scale representation learning by exploiting the inherent multi-resolution temporal characteristics of time series.
For instance, MSDformer~\citep{msdformer} introduces a multi-scale VQ-VAE as the tokenizer to better capture temporal dynamics at different temporal resolutions.
Inspired by scale-wise autoregressive modeling~\citep{VAR}, TimeMAR~\citep{timemar} introduces a next-scale prediction pipeline for time series generation.
In contrast to existing two-stage deep generative models, ORBIT disentangles heterogeneous temporal patterns from a spectral perspective for decoupled representation learning, and further develops a novel autoregressive direction driven by next frequency-band prediction.

\subsection{Autoregressive Modeling for Time Series Generation}
Autoregressive models were initially widely adopted in computer vision~\citep{visionsurvey,followpose,ma2026group}, natural language processing~\citep{languagesurvey}, and audio generation~\citep{audiogeneration}.
This has also sparked considerable interest among researchers and practitioners in the field of time series generation, motivating extensive efforts to adapt them for generating high-quality and high-fidelity time series samples.
For example, GPT-style unidirectional autoregressive models are adopted by methods such as SDformer~\citep{sdformer}, MSDformer~\citep{msdformer}, and Polarformer~\citep{polarformer}.
Along another line of research, BERT-style bidirectional autoregressive models are also explored, like HALO~\citep{halo}, TimeVQVAE~\citep{timevqvae} and SDformer-m~\citep{sdformer}.
More notably, diffusion models can be viewed as a generalized autoregressive paradigm, where samples are generated through iterative denoising and information recovery~\citep{ddpm}.
Diffusion-TS~\citep{diffusionts} incorporates seasonal-trend
decomposition into a Transformer-based denoising network for interpretable time series generation.
TimeDP~\citep{timedp} enables multi-domain time series generation
through a conditional diffusion model guided by prototype-based
domain prompts extracted from a few target-domain examples,
supporting unseen-domain generation without additional fine-tuning.
Different from existing autoregressive models, ORBIT explicitly aligns hierarchical discrete representations with frequency bands and develops an innovative next frequency-band autoregressive modeling paradigm, progressing from low- to high-frequency bands to coordinate the generation of heterogeneous temporal patterns.

\section{Preliminaries}
\subsection{Definitions}
\paragraph{\textbf{Definition 1 Multivariate Time Series}} 
For the given multivariate time series $\mathbf{X}\in\mathbb{R}^{C\times O}$,
where $C$ represents the number of channels, and $O$ is the number of observed time steps.
ORBIT adopts a channel-independent strategy~\citep{patchtst} for robust modeling. 
To this end, for simplicity, we omit the channel dimension in the following formulation.

\paragraph{\textbf{Definition 2 VQ-VAE}}
Let the codebook be $\mathbf{E}=\{\mathbf{e}_{k}\}_{1}^{K}$, where each $\mathbf{e}_{k}\in\mathbb{R}^{d}$ is a discrete vector (code).
Given an input sample $\mathbf{X}$, the encoder $f_{\theta}(\cdot)$ produces a continuous latent temporal representation $\mathbf{z}=f_{\theta}(\mathbf{X})$.
Vector quantization maps \(\mathbf{z}\) to its nearest codebook vector in \(\mathbf{E}\) through a nearest-neighbor lookup:
\begin{equation}
\mathbf{z}_{q}=\mathbf{e}_{k^{*}},\qquad
{k^{*}}=\text{Lookup}(\mathbf{z}; \mathbf{E}),
\end{equation}
where \(\operatorname{Lookup}(\cdot;\cdot)\) denotes the function that retrieves the codebook vector in \(\mathbf{E}\) closest to \(\mathbf{z}\).

The quantized vector \(\mathbf{z}_q\) is subsequently passed through the decoder \(g_{\phi}(\cdot)\) to reconstruct the raw input:
\begin{equation}
\hat{\mathcal{X}} = g_{\phi}(\mathbf{z}_q).
\end{equation}
In practice, the encoder outputs a sequence of latent temporal representations that are independently quantized at each position, yielding a sequence of discrete tokens.
An autoregressive model can then be trained over these discrete tokens to model their joint probability distribution, thereby enabling sequential generation.

\subsection{Problem Statement}
Given a training set \(\mathcal{D}=\{X^{(i)}\}_{i=1}^{N}\) consisting of \(N\) time series samples, time series generation aims to learn a parameterized distribution \(p_{\theta}(X)\) that approximates the underlying data distribution \(p_{\mathrm{data}}(X)\). The model parameters \(\theta\) are estimated from the entire training set \(\mathcal{D}\) by optimizing a suitable training objective, such as a likelihood- or reconstruction-based objective. Once trained, the model can generate synthetic sequences \(\hat{X}\sim p_{\theta}(\cdot)\) whose statistical properties and temporal dynamics resemble those of the real input data.

\section{Methodologies}
\subsection{Overall Framework}

\begin{figure*}[t]
    \centering
    \includegraphics[width=1.\linewidth]{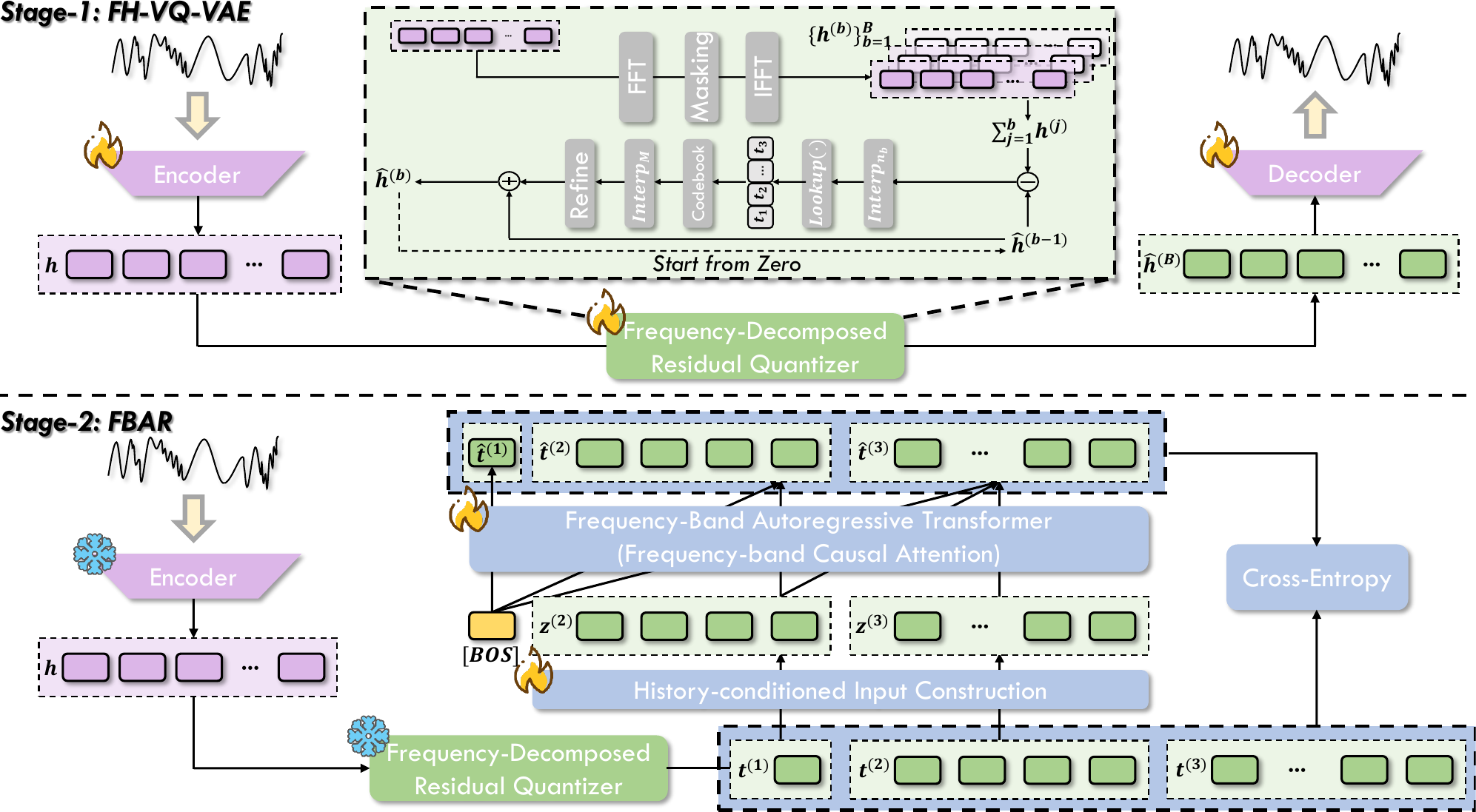}
    \caption{\icon The overview of our proposed ORBIT.}
    \label{fig:3}
\end{figure*}
To achieve high-quality time series generation, we introduce {ORBIT}, a unified deep generative framework whose overall architecture is illustrated in Figure~\ref{fig:3}.
Specifically, \emph{in the first stage}, we train
a Frequency-Hierarchical VQ-VAE (FH-VQ-VAE).
The \textbf{Encoder} compresses the input time series into continuous latent temporal representations.
We then apply \textbf{Frequency-Decomposed Residual Quantizer} to construct hierarchical discrete tokens from low- to high-frequency bands, with each level incorporating a current frequency band while refining the quantization errors left by preceding levels.
The resulting quantized representations are mapped to a common temporal resolution and aggregated before being fed into the \textbf{Decoder} to reconstruct the original time series.

\emph{In the second stage}, we freeze the
FH-VQ-VAE and train a {Frequency-Band Autoregressive Transformer (FBAR)} following the proposed next frequency-band prediction paradigm.
The \textbf{History-conditioned Input Construction} module reconstructs and accumulates preceding frequency bands from their codebook indices, then applies interpolation,
projection, and position and frequency-band embeddings to construct the Transformer inputs.
During training, these inputs are constructed from
ground-truth indices, and FBAR is optimized using
cross-entropy under a \textbf{Frequency-band Causal Attention} to predict
the indices of each subsequent frequency band.
During inference, FBAR starts from a conditional or
unconditional and generates frequency bands sequentially from low to high, while sampling all indices within each band in parallel.
Each generated band is converted into a refined reconstruction contribution through the frozen quantizer and added to the cumulative latent representation, which is used to construct the next frequency-band input.
After all bands are generated, the final accumulated
representation is fed into the frozen decoder of the FH-VQ-VAE to synthesize the time series.

\subsection{Stage 1: Frequency-Hierarchical VQ-VAE}

\textbf{{Encoder.}} 
For the encoder of FH-VQ-VAE, following conventional practice, we progressively compress the input time series into low-dimensional latent representations $\mathbf{h}\in\mathbb{R}^{M\times d}$, where $M$ denotes the number of latent representations, and $s$ is the compression ratio. 
Formally, it can be expressed as follows:
\begin{equation}
    \mathbf{h}=\text{Encoder}(\mathbf{X}).
\end{equation}
Architecturally, the encoder employs 1D convolutional layers with residual blocks, while the decoder symmetrically mirrors the encoder architecture.

\noindent\textbf{Frequency-Decomposed Residual Quantizer.}
After obtaining the encoded latent representations, we leverage frequency-decomposed residual quantizer to obtain hierarchical discrete representations.
This component constitutes the cornerstone of both disentangled representation learning for heterogeneous temporal patterns and the frequency-band autoregressive generation direction.

We first use the Fast Fourier Transform (FFT) operator to decompose the continuous latent representations \(\mathbf{h}\), obtaining its corresponding frequency-domain representations $\mathbf{h}_{f}$:
\begin{equation}
    \mathbf{h}_{f}=\text{FFT}(\mathbf{h}).
\end{equation}
Subsequently, we introduce a frequency-band mask to partition \(\mathbf{h}_f\) into multiple non-overlapping frequency bands.
This operation disentangles heterogeneous temporal patterns by separating them in the spectral domain.
Formally, this process can be expressed as follows:
\begin{equation}
\mathbf{h}_{f}^{(b)}=\mathbf{M}_{b}\odot\mathbf{h}_{f},\quad b=1,\ldots,B,
\end{equation}
where \(\mathbf{M}_{b}\) denotes the frequency-band mask corresponding to the \(b\)-th frequency band.
We then apply the Inverse Fast Fourier Transform (IFFT) to each frequency-band representation, transforming them back to the time domain for subsequent residual quantization:
\begin{equation}
\mathbf{h}^{(b)}
=
\text{IFFT}\!\left(\mathbf{h}_{f}^{(b)}\right),
\qquad b=1,\ldots,B.
\end{equation}

After obtaining \(\{\mathbf{h}^{(b)}\}_{b=1}^{B}\), we apply residual quantization to progressively construct hierarchical discrete representations:
\begin{equation}
\mathbf r^{(b)}
=
\sum_{j=1}^{b}\mathbf h^{(j)}
-
\hat{\mathbf h}^{(b-1)},
\qquad
\hat{\mathbf h}^{(0)}=\mathbf 0,
\end{equation}
where $\mathbf r^{(b)}\in\mathbb{R}^{M\times d}$ denotes the quantization residual of the $b$-th frequency-band level.
Considering that low-frequency components reflect slowly varying structures while high-frequency components contain rapid temporal variations, we adopt frequency-dependent temporal quantization resolutions, allocating fewer tokens to lower-frequency bands and more tokens to higher-frequency bands:
\begin{equation}
    \mathbf{t}^{(b)}=\text{Lookup}(\text{Interp}_{n_b}(\mathbf{r}^{(b)});\mathbf{E}),
\end{equation}
where $\mathbf{t}^{(b)} \in \{1,\ldots,K\}^{n_b}$ denotes the discrete codebook index sequence at the $b$-th frequency-band level, $n_b$ is the number of tokens allocated to this stage, and $\text{Interp}_{n_b}(\cdot)$ resizes the residual representation to $n_b$ temporal positions. $\mathbf{E} \in \mathbb{R}^{K\times d}$ denotes the shared learnable codebook, and $\text{Lookup}(\cdot;\mathbf{E})$ assigns each interpolated feature vector to the index of its closest codebook entry.
Through codebook lookup and resolution restoration, we obtain the continuous reconstruction contribution at the \(b\)-th stage, \(\mathbf{g}^{(b)}\in\mathbb{R}^{M\times d}\), and update the cumulative reconstruction as

\begin{equation}
\widehat{\mathbf{h}}^{(b)}
=
\widehat{\mathbf{h}}^{(b-1)}
+
\mathbf{g}^{(b)}.
\end{equation}

By allocating band-specific token budgets, the above procedure constructs hierarchical representations tailored to heterogeneous temporal patterns.
Moreover, inspired by~\citep{VAR}, we introduce lightweight refinement modules at each quantization stage to adjust the interpolated codeword vectors before accumulating them for latent reconstruction.
For clarity, we provide a detailed pseudo-algorithm in Algorithm~\ref{alg:1}.

\noindent\textbf{Decoder.}
Through the decoder, we reconstruct the final latent representation \(\widehat{\mathbf{h}}^{(B)}\) back into the time series space:
\begin{equation}
\widehat{\mathbf{X}}
=
\text{Decoder}\!\left(\widehat{\mathbf{h}}^{(B)}\right),\quad \widehat{\mathbf{X}}\in\mathbb{R}^{L}.
\end{equation}
The first stage is trained end-to-end by minimizing the reconstruction loss between the input time series and its reconstruction, while the discrete codebooks are jointly optimized using the straight-through estimator~\citep{vqvae}.
Consequently, the overall learning objective is defined as follows:
\begin{equation}
\begin{aligned}
\mathcal{L}_{\mathrm{Stage\text{-}1}}
=&~
\underbrace{
\frac{1}{L}
\left\|
\mathbf{X}-\widehat{\mathbf{X}}
\right\|_{2}^{2}
}_{\mathcal{L}_{\mathrm{rec}}}
\\
&+
\lambda_{\mathrm{cb}}
\underbrace{
\frac{1}{BMd}
\sum_{b=1}^{B}
\left\|
\operatorname{sg}\!\left[
\sum_{j=1}^{b}\mathbf{h}^{(j)}
\right]
-
\widehat{\mathbf{h}}^{(b)}
\right\|_{F}^{2}
}_{\mathcal{L}_{\mathrm{cb}}}
\\
&+
\lambda_{\mathrm{com}}
\underbrace{
\frac{1}{Md}
\sum_{b=1}^{B}
\left\|
\sum_{j=1}^{b}\mathbf{h}^{(j)}
-
\operatorname{sg}\!\left[
\widehat{\mathbf{h}}^{(b)}
\right]
\right\|_{F}^{2}
}_{\mathcal{L}_{\mathrm{com}}},
\end{aligned}
\end{equation}
where $\operatorname{sg}[\cdot]$ denotes the stop-gradient
operator, and $\|\cdot\|_F$ denotes the Frobenius norm.
$\sum_{j=1}^{b}\mathbf{h}^{(j)}$ is the cumulative latent
target up to the $b$-th frequency band, while
$\widehat{\mathbf{h}}^{(b)}$ is its corresponding cumulative
reconstruction obtained through codebook lookup,
interpolation, refinement, and accumulation.
$\lambda_{\mathrm{cb}}$ and $\lambda_{\mathrm{com}}$
control the contributions of the codebook and commitment
losses, respectively.



\begin{algorithm}[h]

\caption{\centering Frequency-Decomposed Residual Quantizer\label{alg:1}}

\begin{algorithmic}[1]
\Require Encoded latent representation
         $\mathbf{h}\in\mathbb{R}^{M\times d}$;
         frequency-band masks $\{\mathbf{M}_b\}_{b=1}^{B}$;
         token budgets $\{n_b\}_{b=1}^{B}$;
         shared codebook $\mathbf{E}\in\mathbb{R}^{K\times d}$;
         refinement modules $\{\phi_b\}_{b=1}^{B}$.
\Ensure Discrete codebook index sequences
        $\{\mathbf{t}^{(b)}\}_{b=1}^{B}$
        and reconstructed latent representation
        $\widehat{\mathbf{h}}^{(B)}$.

\State $\mathbf{h}_f \gets \operatorname{FFT}(\mathbf{h})$
       \Comment{Along the temporal dimension}

\For{$b=1,\ldots,B$}
    \State $\mathbf{h}_f^{(b)}
           \gets \mathbf{M}_b\odot\mathbf{h}_f$
    \State $\mathbf{h}^{(b)}
           \gets \operatorname{IFFT}(\mathbf{h}_f^{(b)})$
\EndFor

\State $\widehat{\mathbf{h}}^{(0)}\gets\mathbf{0}$

\For{$b=1,\ldots,B$}
    \State $\mathbf{r}^{(b)}
           \gets \displaystyle\sum_{j=1}^{b}\mathbf{h}^{(j)}
           -\widehat{\mathbf{h}}^{(b-1)}$
           \Comment{Construct the stage residual}

    \State $\mathbf{t}^{(b)}
           \gets \operatorname{Lookup}
           \bigl(\operatorname{Interp}_{n_b}
           (\mathbf{r}^{(b)});\mathbf{E}\bigr)$
           \Comment{Obtain discrete indices}

    \State $\mathbf{g}^{(b)}
           \gets \phi_b\!\left(
           \operatorname{Interp}_{M}
           \bigl(\mathbf{E}[\mathbf{t}^{(b)}]\bigr)
           \right)$
           \Comment{Restore resolution and refine}

    \State $\widehat{\mathbf{h}}^{(b)}
           \gets \widehat{\mathbf{h}}^{(b-1)}
           +\mathbf{g}^{(b)}$
           \Comment{Accumulate the reconstruction}
\EndFor

\State \Return
       $\{\mathbf{t}^{(b)}\}_{b=1}^{B},
       \widehat{\mathbf{h}}^{(B)}$
\end{algorithmic}
\end{algorithm}

\subsection{Stage 2: Frequency-Band Autoregressive Transformer}

Building on the hierarchical discrete representations learned
by FH-VQ-VAE, we introduce a Frequency-Band Autoregressive
Transformer (FBAR) to model dependencies across frequency bands.
During this stage, FH-VQ-VAE is frozen, and FBAR is trained to
predict the next frequency-band conditioned on preceding bands.
This establishes a progressive autoregressive direction from low-frequency structures to higher-frequency details.

\begin{algorithm}[!h]
\caption{\centering Next Frequency-Band Autoregressive Prediction}
\label{alg:2}

\begin{algorithmic}[1]
\small
\Require Trained FBAR;
         frozen codebook $\mathbf{E}$,
         refinement modules ${\phi_b}{b=1}^{B}$,
         and decoder;
         token budgets ${n_b}{b=1}^{B}$;
         latent length $M$;
         class or null condition $\texttt{[BOS]}$;
         sampling temperature $\tau>0$.
\Ensure Generated time series $\widehat{\mathbf{X}}$.
\State $\widehat{\mathbf{h}}^{(0)}\gets\mathbf{0}$
\State $\mathbf{Z}\gets[,]$
       \Comment{Initialize the input prefix}
\For{$b=1,\ldots,B$}
    \If{$b=1$}
        \State $\mathbf{z}i^{(1)}
               \gets \mathbf{s}i(\texttt{[BOS]})
               +\mathbf{e}{\mathrm{pos},i}^{(1)}
               +\mathbf{e}{\mathrm{band}}^{(1)},
               \quad i=1,\ldots,n_1$
               \Comment{Construct BOS inputs}
    \Else
        \State $\mathbf{u}^{(b)}
               \gets \operatorname{Interp}_{n_b}
               !\left(\widehat{\mathbf{h}}^{(b-1)}\right)$
       \State $\mathbf{z}_i^{(b)}
           \gets \operatorname{Proj}
           \!\left(\mathbf{u}_i^{(b)}\right)
           +\mathbf{e}_{\mathrm{pos},i}^{(b)}
           +\mathbf{e}_{\mathrm{band}}^{(b)},
           \quad i=1,\ldots,n_b$
           \Comment{Construct history-conditioned inputs}
\EndIf

\State $\mathbf{Z}
       \gets \operatorname{Concat}
       \!\left(\mathbf{Z},\mathbf{z}^{(b)}\right)$

\State $\ell_b\gets\displaystyle\sum_{j=1}^{b}n_j$
\State $\mathbf{A}^{(b)}
       \gets\mathbf{A}_{1:\ell_b,\,1:\ell_b}$
       \Comment{Restrict the block-wise causal mask}

\State $\mathbf{Y}
       \gets\operatorname{Transformer}
       \!\left(\mathbf{Z};\mathbf{A}^{(b)}\right)$

\State $\mathbf{Y}^{(b)}
       \gets\mathbf{Y}_{\ell_b-n_b+1:\ell_b,\,:}$
       \Comment{Select current-band hidden states}

\State $\mathbf{P}^{(b)}
       \gets\operatorname{Softmax}
       \!\left(
       \operatorname{Head}
       \!\left(\operatorname{Norm}
       \!\left(\mathbf{Y}^{(b)}\right)\right)/\tau
       \right)$
        
\State $\widehat{t}_i^{(b)}
       \sim\operatorname{Categorical}
       \!\left(\mathbf{P}_i^{(b)}\right),
       \quad i=1,\ldots,n_b$
       \Comment{Sample all indices in parallel (\underline{sampling temperature is fixed at 1.0})}

\State $\mathbf{v}^{(b)}
       \gets\mathbf{E}
       [\widehat{\mathbf{t}}^{(b)}]$
       \Comment{Retrieve codeword vectors}

\State $\widetilde{\mathbf{v}}^{(b)}
       \gets\operatorname{Interp}_{M}
       \!\left(\mathbf{v}^{(b)}\right)$

\State $\mathbf{g}^{(b)}
       \gets\phi_b
       \!\left(\widetilde{\mathbf{v}}^{(b)}\right)$

\State $\widehat{\mathbf{h}}^{(b)}
       \gets\widehat{\mathbf{h}}^{(b-1)}
       +\mathbf{g}^{(b)}$
       \Comment{Accumulate the reconstruction}

\EndFor
\State $\widehat{\mathbf{X}}
       \gets\operatorname{Decoder}
       !\left(\widehat{\mathbf{h}}^{(B)}\right)$
\State \Return $\widehat{\mathbf{X}}$
\end{algorithmic}
\end{algorithm}
Given the discrete codebook index sequences
$\{\mathbf{t}^{(b)}\}_{b=1}^{B}$, FBAR factorizes their joint
distribution as
\begin{equation}
p_{\theta}\!\left(
\mathbf{t}^{(1)},\ldots,\mathbf{t}^{(B)}\mid \texttt{[BOS]}
\right)
=
\prod_{b=1}^{B}
p_{\theta}\!\left(
\mathbf{t}^{(b)}\mid\mathbf{t}^{(<b)},\texttt{[BOS]}
\right),
\end{equation}
where $\mathbf{t}^{(<b)}$ denotes the index sequences of all
preceding frequency bands, and $\texttt{[BOS]}$ denotes an optional class
condition. Within each band, the token distributions are
conditionally factorized and predicted in parallel:
\begin{equation}
p_{\theta}\!\left(
\mathbf{t}^{(b)}\mid\mathbf{t}^{(<b)},\texttt{[BOS]}
\right)
=
\prod_{i=1}^{n_b}
p_{\theta}\!\left(
t_i^{(b)}\mid\mathbf{t}^{(<b)},\texttt{[BOS]}
\right).
\end{equation}

\noindent\textbf{History-conditioned Input Construction.}
Rather than directly shifting the discrete index sequence~\citep{sdformer,msdformer,timemar},
we construct the input of each subsequent stage from the
cumulative latent reconstruction of preceding bands.
Specifically, for $b\geq 2$, the history representation is
resized to the token length of the current band:
\begin{equation}
\mathbf{u}^{(b)}
=
\operatorname{Interp}_{n_b}
\!\left(\widehat{\mathbf{h}}^{(b-1)}\right)
\in\mathbb{R}^{n_b\times d}.
\end{equation}
Here, $\widehat{\mathbf{h}}^{(b-1)}$ is obtained through
codebook lookup, resolution restoration, refinement, and
accumulation, following the procedure introduced in Stage~1.
The resulting history features are projected into the
Transformer embedding space and augmented with position and
frequency-band embeddings:
\begin{equation}
\mathbf{z}_{i}^{(b)}
=
\begin{cases}
\mathbf{s}_{i}(\texttt{[BOS]})
+\mathbf{e}_{\mathrm{pos},i}^{(1)}
+\mathbf{e}_{\mathrm{band}}^{(1)},
& b=1,
\\[2pt]
\text{Proj}\!\left(\mathbf{u}_{i}^{(b)}\right)
+\mathbf{e}_{\mathrm{pos},i}^{(b)}
+\mathbf{e}_{\mathrm{band}}^{(b)},
& b\geq 2,
\end{cases}
\end{equation}
where $\text{Proj}(\cdot)$ maps the latent features from
$d$ dimensions to the Transformer embedding dimension $D$.
$\mathbf{s}_{i}(c)$ combines the class embedding with a
learnable start-position embedding, while
$\mathbf{e}_{\mathrm{pos},i}^{(b)}$ and
$\mathbf{e}_{\mathrm{band}}^{(b)}$ encode position and
frequency-band identity, respectively.
For unconditional generation, the class condition is replaced
with a learned null condition (classifier-free guidance).

\noindent\textbf{Frequency-band Causal Attention.}
We concatenate the stage inputs and apply a block-wise causal
mask to organize attention along the frequency-band axis.
Let $b(i)$ denote the frequency-band stage associated with
the $i$-th input position. The additive attention mask is
defined as
\begin{equation}
A_{ij}
=
\begin{cases}
0, & b(j)\leq b(i),\\
-\infty, & b(j)>b(i).
\end{cases}
\end{equation}
The mask permits attention to positions within the same
stage and preceding stages while blocking future stages.
Importantly, the inputs of the current stage are constructed
exclusively from preceding bands, allowing interactions
among history-conditioned positions without exposing the
current target indices.
The resulting hidden states are normalized and passed
through a classification head to obtain a distribution
over the $K$ codebook entries at each position.

\noindent\textbf{Training and Inference.}
During training, teacher forcing~\citep{ntp} constructs each stage input
using the preceding ground-truth index sequences.
All stages are evaluated in a single forward pass under the
block-wise causal mask. FBAR is optimized using token-level
cross-entropy:
\begin{equation}
\mathcal{L}_{\mathrm{Stage\text{-}2}}
=
-\frac{1}{\sum_{b=1}^{B}n_b}
\sum_{b=1}^{B}
\sum_{i=1}^{n_b}
\log
p_{\theta}\!\left(
t_i^{(b)}\mid\mathbf{t}^{(<b)},\texttt{[BOS]}
\right).
\end{equation}

During inference, FBAR starts from the class-conditioned
or unconditional start embeddings and proceeds sequentially
from low- to high-frequency bands.
At each level, all $n_b$ indices are sampled in parallel
from the predicted categorical distributions.
The sampled indices are converted into a reconstruction
contribution using the frozen codebook and refinement modules,
and added to the cumulative latent reconstruction.
This reconstruction is then resized to construct the input
for the next frequency-band.
After all $B$ stages, the final accumulated latent
representation is fed into the frozen decoder:
\begin{equation}
\widehat{\mathbf{X}}
=
\operatorname{Decoder}
\!\left(\widehat{\mathbf{h}}^{(B)}\right).
\end{equation}
Thus, generation requires $B$ sequential band-level steps,
with token positions within each band generated in parallel.
Benefiting from this frequency-band-level parallel decoding capability, ORBIT also achieves improving inference efficiency.
For clarity, we present the inference procedure in Algorithm~\ref{alg:2}.

\begin{table*}[!ht]
    \centering
    \caption{
        Maximum mean discrepancy (MMD) and Kullback--Leibler divergence (K-L)
        for time series generation at sequence lengths 24, 96, and 168.
        Lower values indicate better performance.
        Results at individual lengths are reported as mean $\pm$ standard deviation.
        Best means are highlighted in red and bold, and second-best results
        are highlighted in blue and underlined.\label{tab:1}}
   
    \vspace{-6pt}
    \setlength{\tabcolsep}{2.8pt}
    \renewcommand{\arraystretch}{0.74}

    \newcommand{\res}[2]{%
        \ensuremath{#1\!\pm\!#2}%
    }
    \newcommand{\best}[2]{%
        \ensuremath{\textcolor{red}{\mathbf{#1}\!\pm\!#2}}%
    }
    \newcommand{\second}[2]{%
        \ensuremath{\textcolor{blue}{\underline{#1\!\pm\!#2}}}%
    }

    \resizebox{\textwidth}{!}{%
    \begin{tabular}{cc|cc|cc|cc|cc|cc|cc|cc|cc}
        \toprule[1.5pt]
        \multirow{2}{*}{Dataset}
        & \multirow{2}{*}{Length}
        & \multicolumn{2}{c|}{ORBIT (\textbf{Ours})}
        & \multicolumn{2}{c|}{TimeMAR~\cite{timemar}}
        & \multicolumn{2}{c|}{TimeDP~\cite{timedp}}
        & \multicolumn{2}{c|}{TimeVQVAE~\cite{timevqvae}}
        & \multicolumn{2}{c|}{TimeVQVAE-C~\cite{timevqvae}}
        & \multicolumn{2}{c|}{TimeVAE~\cite{timevae}}
        & \multicolumn{2}{c|}{GT-GAN~\cite{gtgan}}
        & \multicolumn{2}{c}{TimeGAN~\cite{TIMEGAN}} \\
        & & MMD & K-L & MMD & K-L & MMD & K-L
          & MMD & K-L & MMD & K-L & MMD & K-L
          & MMD & K-L & MMD & K-L \\
        \midrule[0.5pt]

        \multirow{4}{*}{Electricity}
        & 24
        & \best{0.001}{0.000} & \best{0.007}{0.002}
        & \second{0.001}{0.001} & \second{0.010}{0.000}
        & \second{0.001}{0.001} & \res{0.016}{0.015}
        & \res{0.142}{0.017} & \res{0.302}{0.022}
        & \res{0.002}{0.001} & \res{0.023}{0.007}
        & \res{0.980}{0.016} & \res{1.245}{0.018}
        & \res{0.189}{0.085} & \res{0.423}{0.178}
        & \res{0.270}{0.095} & \res{0.333}{0.134} \\
        & 96
        & \best{0.001}{0.000} & \res{0.029}{0.005}
        & \second{0.001}{0.001} & \second{0.026}{0.004}
        & \res{0.002}{0.001} & \best{0.015}{0.017}
        & \res{0.153}{0.023} & \res{0.245}{0.039}
        & \res{0.002}{0.000} & \res{0.031}{0.005}
        & \res{0.648}{0.012} & \res{0.790}{0.018}
        & \res{0.317}{0.311} & \res{0.447}{0.147}
        & \res{0.401}{0.219} & \res{0.552}{0.241} \\
        & 168
        & \best{0.001}{0.000} & \best{0.012}{0.002}
        & \second{0.001}{0.001} & \second{0.012}{0.003}
        & \second{0.001}{0.001} & \res{0.012}{0.016}
        & \res{0.152}{0.024} & \res{0.280}{0.051}
        & \res{0.002}{0.001} & \res{0.027}{0.015}
        & \res{0.577}{0.006} & \res{0.734}{0.023}
        & \res{0.254}{0.166} & \res{0.407}{0.079}
        & \res{0.367}{0.255} & \res{0.488}{0.175} \\
        \midrule[0.5pt]
        & Average
        & \textcolor{red}{\textbf{0.001}} & \textcolor{blue}{\underline{0.016}}
        & \textcolor{blue}{\underline{0.001}} & \textcolor{blue}{\underline{0.016}}
        & 0.001 & \textcolor{red}{\textbf{0.014}}
        & 0.149 & 0.276
        & 0.002 & 0.027
        & 0.735 & 0.923
        & 0.253 & 0.426
        & 0.346 & 0.458 \\
        \midrule[0.5pt]

        \multirow{4}{*}{Solar}
        & 24
        & \best{0.011}{0.001} & \best{0.012}{0.001}
        & \second{0.015}{0.000} & \second{0.013}{0.000}
        & \res{0.023}{0.016} & \res{0.017}{0.009}
        & \res{0.411}{0.004} & \res{0.878}{0.031}
        & \res{0.036}{0.011} & \res{0.120}{0.036}
        & \res{0.321}{0.005} & \res{0.458}{0.016}
        & \res{0.517}{0.027} & \res{0.102}{0.039}
        & \res{0.513}{0.031} & \res{0.410}{0.042} \\
        & 96
        & \second{0.014}{0.000} & \second{0.013}{0.001}
        & \best{0.012}{0.000} & \res{0.014}{0.000}
        & \res{0.019}{0.001} & \best{0.010}{0.002}
        & \res{0.439}{0.021} & \res{0.812}{0.077}
        & \res{0.035}{0.008} & \res{0.151}{0.114}
        & \res{0.369}{0.012} & \res{0.288}{0.015}
        & \res{0.634}{0.119} & \res{0.117}{0.047}
        & \res{0.584}{0.056} & \res{0.922}{0.953} \\
        & 168
        & \second{0.038}{0.001} & \res{0.019}{0.000}
        & \best{0.032}{0.000} & \best{0.012}{0.000}
        & \res{0.041}{0.011} & \second{0.016}{0.005}
        & \res{0.437}{0.020} & \res{0.865}{0.108}
        & \res{0.058}{0.005} & \res{0.234}{0.062}
        & \res{0.353}{0.014} & \res{0.260}{0.016}
        & \res{0.578}{0.039} & \res{0.120}{0.041}
        & \res{0.628}{0.053} & \res{0.612}{0.447} \\
        \midrule[0.5pt]
                & Average
        & \textcolor{blue}{\underline{0.021}} & 0.015
        & \textcolor{red}{\textbf{0.020}} & \textcolor{red}{\textbf{0.013}}
        & 0.028 & \textcolor{blue}{\underline{0.014}}
        & 0.429 & 0.852
        & 0.043 & 0.168
        & 0.348 & 0.335
        & 0.576 & 0.113
        & 0.575 & 0.648 \\
        \midrule[0.5pt]

        \multirow{4}{*}{Wind}
        & 24
        & \best{0.021}{0.003} & \best{0.098}{0.007}
        & \second{0.033}{0.000} & \second{0.126}{0.000}
        & \res{0.042}{0.019} & \res{0.316}{0.057}
        & \res{0.129}{0.008} & \res{0.566}{0.030}
        & \res{0.034}{0.011} & \res{0.374}{0.053}
        & \res{0.419}{0.006} & \res{1.105}{0.019}
        & \res{0.308}{0.015} & \res{0.317}{0.175}
        & \res{0.348}{0.019} & \res{2.378}{0.989} \\
        & 96
        & \second{0.022}{0.002} & \best{0.094}{0.006}
        & \res{0.031}{0.002} & \second{0.122}{0.005}
        & \res{0.029}{0.015} & \res{0.182}{0.029}
        & \res{0.129}{0.016} & \res{0.510}{0.112}
        & \best{0.013}{0.007} & \res{0.284}{0.115}
        & \res{0.179}{0.002} & \res{0.527}{0.009}
        & \res{0.217}{0.134} & \res{0.164}{0.071}
        & \res{0.195}{0.030} & \res{2.288}{1.135} \\
        & 168
        & \res{0.025}{0.003} & \best{0.100}{0.009}
        & \second{0.023}{0.001} & \second{0.107}{0.006}
        & \res{0.025}{0.017} & \res{0.152}{0.034}
        & \res{0.131}{0.014} & \res{0.483}{0.066}
        & \best{0.018}{0.007} & \res{0.183}{0.047}
        & \res{0.170}{0.004} & \res{0.484}{0.015}
        & \res{0.170}{0.040} & \res{0.107}{0.016}
        & \res{0.213}{0.017} & \res{1.924}{1.233} \\
        \midrule[0.5pt]
                & Average
        & \textcolor{blue}{\underline{0.023}} & \textcolor{red}{\textbf{0.097}}
        & 0.029 & \textcolor{blue}{\underline{0.118}}
        & 0.032 & 0.217
        & 0.130 & 0.520
        & \textcolor{red}{\textbf{0.022}} & 0.280
        & 0.256 & 0.705
        & 0.232 & 0.196
        & 0.252 & 2.197 \\
        \midrule[0.5pt]

        \multirow{4}{*}{Traffic}
        & 24
        & \second{0.004}{0.000} & \best{0.007}{0.002}
        & \best{0.003}{0.000} & \second{0.009}{0.000}
        & \res{0.009}{0.005} & \res{0.014}{0.005}
        & \res{0.189}{0.009} & \res{0.219}{0.016}
        & \res{0.014}{0.004} & \res{0.013}{0.005}
        & \res{0.146}{0.005} & \res{0.446}{0.116}
        & \res{0.522}{0.072} & \res{1.211}{0.286}
        & \res{0.466}{0.071} & \res{1.061}{0.177} \\
        & 96
        & \best{0.016}{0.001} & \best{0.010}{0.001}
        & \res{0.023}{0.000} & \res{0.019}{0.000}
        & \second{0.018}{0.005} & \second{0.010}{0.006}
        & \res{0.189}{0.014} & \res{0.192}{0.016}
        & \res{0.027}{0.002} & \res{0.014}{0.004}
        & \res{0.162}{0.005} & \res{0.205}{0.008}
        & \res{0.513}{0.073} & \res{1.198}{0.340}
        & \res{0.456}{0.100} & \res{1.241}{0.757} \\
        & 168
        & \best{0.065}{0.000} & \best{0.008}{0.001}
        & \second{0.068}{0.000} & \res{0.014}{0.001}
        & \res{0.083}{0.034} & \second{0.009}{0.003}
        & \res{0.213}{0.016} & \res{0.178}{0.026}
        & \res{0.089}{0.001} & \res{0.016}{0.003}
        & \res{0.218}{0.007} & \res{0.211}{0.014}
        & \res{0.538}{0.078} & \res{1.409}{0.251}
        & \res{0.567}{0.057} & \res{1.305}{0.320} \\
        \midrule[0.5pt]
                & Average
        & \textcolor{red}{\textbf{0.028}} & \textcolor{red}{\textbf{0.008}}
        & \textcolor{blue}{\underline{0.031}} & 0.014
        & 0.037 & \textcolor{blue}{\underline{0.011}}
        & 0.197 & 0.196
        & 0.043 & 0.014
        & 0.175 & 0.287
        & 0.524 & 1.273
        & 0.496 & 1.202 \\
        \midrule[0.5pt]

        \multirow{4}{*}{Taxi}
        & 24
        & \second{0.021}{0.000} & \res{0.015}{0.001}
        & \best{0.013}{0.000} & \best{0.004}{0.000}
        & \res{0.032}{0.005} & \second{0.006}{0.002}
        & \res{0.036}{0.003} & \res{0.121}{0.015}
        & \res{0.036}{0.004} & \res{0.015}{0.007}
        & \res{0.049}{0.004} & \res{0.263}{0.022}
        & \res{0.243}{0.030} & \res{0.872}{0.148}
        & \res{0.156}{0.046} & \res{0.547}{0.150} \\
        & 96
        & \best{0.054}{0.001} & \second{0.024}{0.001}
        & \res{0.060}{0.001} & \res{0.045}{0.002}
        & \second{0.055}{0.007} & \best{0.010}{0.007}
        & \res{0.089}{0.005} & \res{0.102}{0.017}
        & \res{0.061}{0.002} & \res{0.027}{0.007}
        & \res{0.060}{0.002} & \res{0.121}{0.003}
        & \res{0.328}{0.071} & \res{0.801}{0.270}
        & \res{0.197}{0.057} & \res{0.550}{0.333} \\
        & 168
        & \best{0.089}{0.003} & \second{0.033}{0.002}
        & \res{0.150}{0.002} & \res{0.173}{0.012}
        & \second{0.095}{0.023} & \best{0.011}{0.004}
        & \res{0.128}{0.004} & \res{0.110}{0.026}
        & \res{0.109}{0.014} & \res{0.038}{0.010}
        & \res{0.139}{0.007} & \res{0.110}{0.020}
        & \res{0.319}{0.032} & \res{0.950}{0.197}
        & \res{0.275}{0.054} & \res{0.650}{0.180} \\
        \midrule[0.5pt]
                & Average
        & \textcolor{red}{\textbf{0.055}} & \textcolor{blue}{\underline{0.024}}
        & 0.074 & 0.074
        & \textcolor{blue}{\underline{0.061}} & \textcolor{red}{\textbf{0.009}}
        & 0.084 & 0.111
        & 0.069 & 0.027
        & 0.083 & 0.165
        & 0.297 & 0.874
        & 0.209 & 0.582 \\
        \midrule[0.5pt]

        \multirow{4}{*}{Pedestrian}
        & 24
        & \best{0.033}{0.001} & \best{0.015}{0.001}
        & \second{0.044}{0.000} & \res{0.019}{0.000}
        & \res{0.056}{0.011} & \second{0.016}{0.010}
        & \res{0.087}{0.005} & \res{0.431}{0.017}
        & \res{0.063}{0.004} & \res{0.023}{0.008}
        & \res{0.111}{0.003} & \res{0.211}{0.015}
        & \res{0.121}{0.011} & \res{0.408}{0.105}
        & \res{0.085}{0.022} & \res{0.369}{0.092} \\
        & 96
        & \second{0.042}{0.001} & \second{0.015}{0.000}
        & \res{0.056}{0.001} & \res{0.040}{0.003}
        & \best{0.040}{0.010} & \best{0.012}{0.005}
        & \res{0.066}{0.011} & \res{0.373}{0.027}
        & \res{0.049}{0.005} & \res{0.034}{0.006}
        & \res{0.064}{0.001} & \res{0.074}{0.004}
        & \res{0.159}{0.149} & \res{0.327}{0.148}
        & \res{0.068}{0.025} & \res{0.458}{0.368} \\
        & 168
        & \best{0.044}{0.001} & \second{0.015}{0.001}
        & \res{0.067}{0.001} & \res{0.035}{0.001}
        & \second{0.044}{0.020} & \best{0.014}{0.010}
        & \res{0.067}{0.007} & \res{0.405}{0.051}
        & \res{0.058}{0.002} & \res{0.039}{0.008}
        & \res{0.065}{0.002} & \res{0.065}{0.005}
        & \res{0.112}{0.019} & \res{0.411}{0.096}
        & \res{0.090}{0.030} & \res{0.417}{0.181} \\
        \midrule[0.5pt]
                & Average
        & \textcolor{red}{\textbf{0.040}} & \textcolor{blue}{\underline{0.015}}
        & 0.056 & 0.031
        & \textcolor{blue}{\underline{0.047}} & \textcolor{red}{\textbf{0.014}}
        & 0.073 & 0.403
        & 0.057 & 0.032
        & 0.080 & 0.117
        & 0.131 & 0.382
        & 0.081 & 0.415 \\
        \midrule[0.5pt]

        \multirow{4}{*}{Air}
        & 24
        & \res{0.009}{0.001} & \res{0.033}{0.003}
        & \best{0.001}{0.000} & \best{0.012}{0.000}
        & \second{0.008}{0.004} & \second{0.029}{0.007}
        & \res{0.010}{0.002} & \res{0.108}{0.015}
        & \res{0.017}{0.005} & \res{0.057}{0.006}
        & \res{0.192}{0.010} & \res{0.683}{0.067}
        & \res{0.206}{0.007} & \res{0.923}{0.119}
        & \res{0.146}{0.022} & \res{0.477}{0.154} \\
        & 96
        & \second{0.009}{0.001} & \res{0.030}{0.002}
        & \best{0.008}{0.000} & \best{0.022}{0.001}
        & \res{0.010}{0.006} & \second{0.025}{0.011}
        & \res{0.022}{0.003} & \res{0.061}{0.012}
        & \res{0.036}{0.007} & \res{0.096}{0.019}
        & \res{0.085}{0.011} & \res{0.168}{0.012}
        & \res{0.244}{0.160} & \res{0.510}{0.147}
        & \res{0.103}{0.046} & \res{0.317}{0.102} \\
        & 168
        & \second{0.013}{0.001} & \second{0.041}{0.003}
        & \res{0.031}{0.001} & \res{0.070}{0.005}
        & \best{0.011}{0.003} & \best{0.027}{0.016}
        & \res{0.028}{0.002} & \res{0.054}{0.012}
        & \res{0.041}{0.008} & \res{0.093}{0.025}
        & \res{0.089}{0.016} & \res{0.164}{0.012}
        & \res{0.211}{0.041} & \res{0.578}{0.049}
        & \res{0.120}{0.045} & \res{0.348}{0.093} \\
        \midrule[0.5pt]
                & Average
        & \textcolor{red}{\textbf{0.010}} & \textcolor{blue}{\underline{0.035}}
        & 0.013 & \textcolor{blue}{\underline{0.035}}
        & \textcolor{blue}{\underline{0.010}} & \textcolor{red}{\textbf{0.027}}
        & 0.020 & 0.074
        & 0.031 & 0.082
        & 0.122 & 0.338
        & 0.220 & 0.670
        & 0.123 & 0.381 \\
        \midrule[0.5pt]

        \multirow{4}{*}{Temperature}
        & 24
        & \second{0.085}{0.003} & \second{0.200}{0.010}
        & \best{0.058}{0.000} & \best{0.142}{0.000}
        & \res{0.086}{0.015} & \res{0.240}{0.029}
        & \res{0.215}{0.018} & \res{0.764}{0.076}
        & \res{0.111}{0.019} & \res{0.357}{0.055}
        & \res{1.003}{0.012} & \res{1.288}{0.145}
        & \res{0.801}{0.015} & \res{1.419}{0.461}
        & \res{0.801}{0.048} & \res{7.672}{2.287} \\
        & 96
        & \best{0.152}{0.002} & \best{0.354}{0.007}
        & \second{0.177}{0.005} & \second{0.388}{0.013}
        & \res{0.178}{0.047} & \res{0.392}{0.097}
        & \res{0.413}{0.034} & \res{1.135}{0.110}
        & \res{0.205}{0.033} & \res{0.567}{0.050}
        & \res{0.958}{0.012} & \res{1.737}{0.068}
        & \res{0.859}{0.109} & \res{5.830}{4.888}
        & \res{0.827}{0.092} & \res{11.298}{1.212} \\
        & 168
        & \best{0.212}{0.003} & \best{0.167}{0.002}
        & \res{0.235}{0.005} & \res{0.220}{0.011}
        & \second{0.219}{0.022} & \second{0.171}{0.073}
        & \res{0.323}{0.008} & \res{0.735}{0.066}
        & \res{0.259}{0.043} & \res{0.379}{0.110}
        & \res{1.002}{0.014} & \res{2.183}{0.110}
        & \res{0.809}{0.081} & \res{3.174}{2.685}
        & \res{0.926}{0.042} & \res{8.892}{2.681} \\
        \midrule[0.5pt]
                & Average
        & \textcolor{red}{\textbf{0.150}} & \textcolor{red}{\textbf{0.240}}
        & \textcolor{blue}{\underline{0.157}} & \textcolor{blue}{\underline{0.250}}
        & 0.161 & 0.268
        & 0.317 & 0.878
        & 0.192 & 0.434
        & 0.988 & 1.736
        & 0.823 & 3.474
        & 0.851 & 9.287 \\
        \midrule[0.5pt]

        \multirow{4}{*}{Rain}
        & 24
        & \second{0.025}{0.001} & \best{0.014}{0.003}
        & \res{0.026}{0.000} & \res{0.053}{0.000}
        & \best{0.014}{0.010} & \second{0.037}{0.019}
        & \res{0.036}{0.006} & \res{0.077}{0.020}
        & \res{0.037}{0.004} & \res{0.089}{0.037}
        & \res{0.737}{0.048} & \res{0.987}{0.073}
        & \res{0.085}{0.052} & \res{0.468}{0.158}
        & \res{0.182}{0.096} & \res{0.356}{0.089} \\
        & 96
        & \second{0.056}{0.001} & \res{0.092}{0.003}
        & \second{0.056}{0.001} & \res{0.096}{0.002}
        & \best{0.044}{0.027} & \best{0.022}{0.009}
        & \res{0.068}{0.008} & \second{0.055}{0.021}
        & \res{0.072}{0.001} & \res{0.091}{0.010}
        & \res{0.327}{0.020} & \res{0.237}{0.025}
        & \res{0.166}{0.203} & \res{0.481}{0.146}
        & \res{0.285}{0.199} & \res{0.549}{0.257} \\
        & 168
        & \best{0.048}{0.002} & \res{0.053}{0.002}
        & \res{0.074}{0.002} & \res{0.081}{0.002}
        & \second{0.057}{0.039} & \best{0.013}{0.012}
        & \res{0.074}{0.007} & \second{0.047}{0.018}
        & \res{0.080}{0.004} & \res{0.065}{0.018}
        & \res{0.292}{0.019} & \res{0.160}{0.022}
        & \res{0.111}{0.109} & \res{0.432}{0.099}
        & \res{0.329}{0.285} & \res{0.506}{0.174} \\
        \midrule[0.5pt]
                & Average
        & \textcolor{blue}{\underline{0.043}} & \textcolor{blue}{\underline{0.053}}
        & 0.052 & 0.077
        & \textcolor{red}{\textbf{0.038}} & \textcolor{red}{\textbf{0.024}}
        & 0.059 & 0.060
        & 0.063 & 0.082
        & 0.452 & 0.461
        & 0.121 & 0.460
        & 0.265 & 0.470 \\
        \midrule[0.5pt]

        \multirow{4}{*}{NN5}
        & 24
        & \second{0.078}{0.001} & \best{0.060}{0.003}
        & \res{0.080}{0.000} & \res{0.069}{0.000}
        & \best{0.072}{0.015} & \res{0.074}{0.028}
        & \res{0.162}{0.011} & \res{0.904}{0.175}
        & \res{0.086}{0.004} & \second{0.063}{0.010}
        & \res{0.708}{0.037} & \res{1.434}{0.117}
        & \res{0.507}{0.078} & \res{1.154}{0.288}
        & \res{0.704}{0.028} & \res{2.273}{0.231} \\
        & 96
        & \second{0.213}{0.002} & \best{0.035}{0.003}
        & \best{0.212}{0.002} & \res{0.071}{0.002}
        & \res{0.234}{0.047} & \second{0.069}{0.033}
        & \res{0.321}{0.008} & \res{1.038}{0.168}
        & \res{0.241}{0.011} & \res{0.103}{0.072}
        & \res{0.875}{0.053} & \res{1.537}{0.224}
        & \res{0.706}{0.116} & \res{2.709}{2.080}
        & \res{0.888}{0.091} & \res{5.829}{6.120} \\
        & 168
        & \best{0.158}{0.001} & \res{0.127}{0.002}
        & \second{0.161}{0.002} & \second{0.094}{0.004}
        & \res{0.164}{0.010} & \best{0.054}{0.014}
        & \res{0.327}{0.012} & \res{1.063}{0.274}
        & \res{0.243}{0.041} & \res{0.220}{0.151}
        & \res{0.821}{0.061} & \res{1.337}{0.220}
        & \res{0.632}{0.074} & \res{1.386}{0.520}
        & \res{0.874}{0.088} & \res{4.928}{4.112} \\
        \midrule[0.5pt]
                & Average
        & \textcolor{red}{\textbf{0.150}} & \textcolor{blue}{\underline{0.074}}
        & \textcolor{blue}{\underline{0.151}} & {{0.078}}
        & 0.157 & \textcolor{red}{\textbf{0.066}}
        & 0.270 & 1.002
        & 0.190 & 0.129
        & 0.801 & 1.436
        & 0.615 & 1.750
        & 0.822 & 4.343 \\
        \midrule[0.5pt]

        \multirow{4}{*}{Fred-MD}
        & 24
        & \best{0.001}{0.000} & \best{0.258}{0.006}
        & \second{0.002}{0.000} & \second{0.316}{0.000}
        & \res{0.002}{0.001} & \res{0.515}{0.174}
        & \res{0.003}{0.001} & \res{0.533}{0.067}
        & \res{0.003}{0.000} & \res{0.574}{0.168}
        & \res{0.187}{0.014} & \res{0.701}{0.091}
        & \res{0.822}{0.139} & \res{2.161}{0.483}
        & \res{1.113}{0.048} & \res{3.356}{0.639} \\
        & 96
        & \second{0.002}{0.000} & \best{0.294}{0.002}
        & \best{0.001}{0.000} & \second{0.328}{0.008}
        & \res{0.005}{0.002} & \res{1.183}{0.280}
        & \res{0.004}{0.001} & \res{0.813}{0.093}
        & \res{0.006}{0.001} & \res{1.240}{0.232}
        & \res{0.046}{0.013} & \res{0.441}{0.063}
        & \res{0.150}{0.184} & \res{0.486}{0.108}
        & \res{0.055}{0.022} & \res{0.628}{0.335} \\
        & 168
        & \best{0.002}{0.000} & \res{0.212}{0.003}
        & \best{0.002}{0.000} & \second{0.207}{0.008}
        & \second{0.002}{0.001} & \best{0.203}{0.035}
        & \res{0.008}{0.002} & \res{0.831}{0.077}
        & \res{0.005}{0.002} & \res{1.118}{0.276}
        & \res{0.059}{0.008} & \res{0.346}{0.041}
        & \res{0.133}{0.102} & \res{0.380}{0.070}
        & \res{0.043}{0.021} & \res{0.512}{0.290} \\
        \midrule[0.5pt]
                & Average
        & \textcolor{red}{\textbf{0.002}} & \textcolor{red}{\textbf{0.255}}
        & \textcolor{red}{\textbf{0.002}} & \textcolor{blue}{\underline{0.284}}
        & \textcolor{blue}{\underline{0.003}} & 0.634
        & 0.005 & 0.726
        & 0.005 & 0.977
        & 0.097 & 0.496
        & 0.368 & 1.009
        & 0.404 & 1.499 \\
        \midrule[0.5pt]

        \multirow{4}{*}{Exchange}
        & 24
        & \best{0.122}{0.002} & \best{2.016}{0.059}
        & \res{0.134}{0.000} & \second{2.098}{0.000}
        & \second{0.127}{0.066} & \res{7.270}{2.080}
        & \res{0.276}{0.026} & \res{4.201}{0.945}
        & \res{0.155}{0.039} & \res{4.561}{0.212}
        & \res{0.662}{0.018} & \res{8.682}{1.334}
        & \res{0.484}{0.074} & \res{3.897}{2.172}
        & \res{0.504}{0.024} & \res{5.482}{2.108} \\
        & 96
        & \best{0.132}{0.003} & \best{2.262}{0.063}
        & \second{0.141}{0.003} & \second{2.660}{0.116}
        & \res{0.343}{0.210} & \res{14.336}{2.652}
        & \res{0.337}{0.056} & \res{3.740}{1.261}
        & \res{0.224}{0.028} & \res{4.408}{0.700}
        & \res{0.592}{0.124} & \res{9.937}{4.322}
        & \res{0.567}{0.114} & \res{10.170}{6.381}
        & \res{0.449}{0.018} & \res{6.445}{2.210} \\
        & 168
        & \second{0.126}{0.002} & \best{1.642}{0.032}
        & \best{0.125}{0.003} & \res{1.887}{0.020}
        & \res{0.151}{0.024} & \second{1.866}{0.132}
        & \res{0.342}{0.050} & \res{5.052}{1.385}
        & \res{0.233}{0.107} & \res{8.475}{3.056}
        & \res{0.543}{0.149} & \res{10.404}{1.434}
        & \res{0.475}{0.116} & \res{7.201}{4.380}
        & \res{0.530}{0.154} & \res{8.861}{3.397} \\
        \midrule[0.5pt]
                & Average
        & \textcolor{red}{\textbf{0.127}} & \textcolor{red}{\textbf{1.973}}
        & \textcolor{blue}{\underline{0.133}} & \textcolor{blue}{\underline{2.215}}
        & 0.207 & 7.824
        & 0.318 & 4.331
        & 0.204 & 5.815
        & 0.599 & 9.674
        & 0.509 & 7.089
        & 0.494 & 6.929 \\
        \midrule[0.5pt]

        \multicolumn{2}{c}{\emph{Wins}}
        & \multicolumn{2}{c}{\textcolor{red}{\textbf{53}}}
        & \multicolumn{2}{c}{19}
        & \multicolumn{2}{c}{\textcolor{blue}{\underline{23}}}
        & \multicolumn{2}{c}{0}
        & \multicolumn{2}{c}{3}
        & \multicolumn{2}{c}{0}
        & \multicolumn{2}{c}{0}
        & \multicolumn{2}{c}{0} \\

        \bottomrule[1.5pt]
    \end{tabular}%
    }
\end{table*}

\section{Experiments}
\subsection{Datasets}
We evaluate ORBIT on 12 datasets across four domains:
energy (Electricity, Solar, Wind), transportation (Traffic,
Taxi, Pedestrian), environment (Air Quality, Temperature,
Rain), and economics (NN5, FRED-MD, Exchange).
These benchmarks are sourced from GluonTS~\citep{gluonts}
and the Monash Time Series Forecasting Repository~\citep{monash}.
Following~\citep{timedp}, we jointly train on the combined
multi-domain dataset and evaluate each domain separately.
Dataset details are provided in Appendix~\ref{app:a}.

\subsection{Baselines}

We compare our method against a set of cutting-edge deep generative baselines, including TimeGAN~\citep{TIMEGAN}, GT-GAN~\citep{gtgan}, TimeVAE~\citep{timevae}, TimeVQVAE~\citep{timevqvae}, TimeVQVAE-C (the class-conditional variant of TimeVQVAE)~\citep{timevqvae}, TimeDP~\citep{timedp}, and TimeMAR ~\citep{timemar}. 
Detailed descriptions of these baselines are provided in Appendix~\ref{app:b}.

\subsection{Implementation Details}

All experiments are conducted on a single NVIDIA RTX 4090 GPU using Python 3.10.2 and PyTorch 2.9.1~\citep{pytorch}. 
Both stages are trained for up to \(1\times10^{6}\) iterations with a batch size of 256 and an initial learning rate of \(1\times10^{-4}\), using a \(1\times10^{3}\)-iteration linear warm-up followed by cosine decay. 
We consider a basic generation length of 24 and further evaluate longer lengths of 96 and 168 to examine the robustness of different models in long-sequence generation.
More implementation details are provided in Appendix~\ref{app:c}.

\subsection{Evaluation Metrics}
To evaluate generation quality, we assess both the distributional fidelity between real and synthetic time series and the preservation of their temporal characteristics. 
Consequently, we adopt two metrics: Maximum Mean Discrepancy (\textbf{MMD}), and Kullback–Leibler (\textbf{K-L}) divergence.
\textbf{MMD} measures the distributional discrepancy between real and synthetic samples in a kernel-induced feature space, while \textbf{K-L} divergence quantifies the discrepancy between their probability distributions. 
For all metrics, lower values indicate better generation quality.

\begin{figure*}[!ht]
    \centering
    \includegraphics[width=0.49\linewidth]{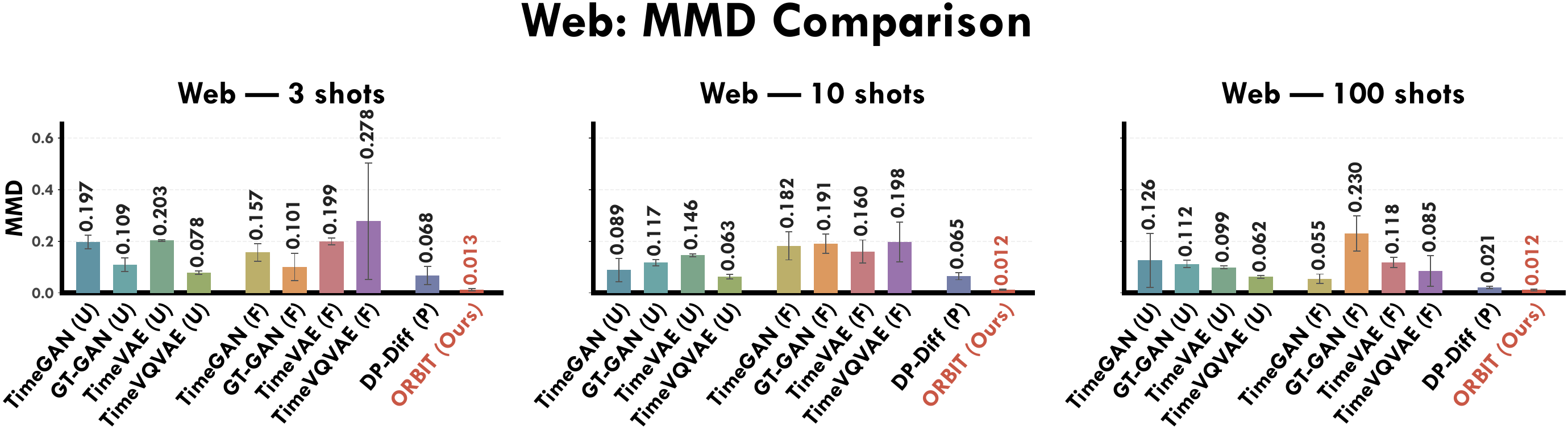} 
    \includegraphics[width=0.49\linewidth]{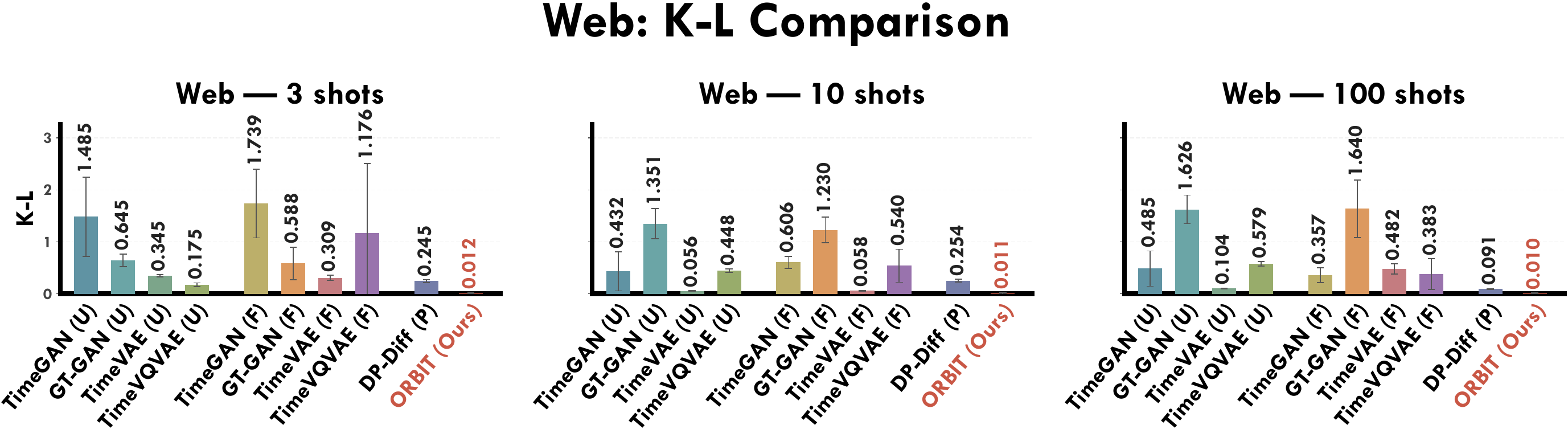} \\
    \vspace{-10pt}
    \caption{MMD and K-L divergence in cross-domain settings of Web datasets for sequence length 168. 
    U, F, and P denote unconditional generation, fine-tuning, and sample prompting, respectively. 
    ORBIT uses CFG to achieve unconditional generation.}
    \label{fig:4}
\end{figure*}
\subsection{Main Results}

\noindent\textbf{In-Domain Generation.}
As shown in Table~\ref{tab:1}, ORBIT achieves state-of-the-art
overall generation performance across 12 datasets and 3
sequence lengths, demonstrating strong distributional fidelity
under both MMD and K-L divergence.
These improvements can be attributed to its
frequency-hierarchical representation learning and next frequency-band
autoregressive direction.
Specifically, FH-VQ-VAE separates frequency components and
progressively refines quantization errors, facilitating the
representation of global temporal structures and fine-grained
details.
FBAR further captures cross-band dependencies through
next frequency-band prediction, enabling coordinated
coarse-to-fine generation.
The combination of these complementary designs contributes
to ORBIT's strong generation performance across diverse
time series datasets.

TimeMAR~\citep{timemar} and TimeDP~\citep{timedp} remain
competitive, yet ORBIT delivers stronger overall performance.
Although TimeMAR learns multi-scale representations, it
retains token-level causal decoding, which may accumulate
prediction errors and compromise generation fidelity.
In contrast, ORBIT generates tokens within each frequency
band in parallel, reducing sequential decoding steps while
modeling cross-band correlations.
TimeDP relies on prototype-based domain prompts, whose
effectiveness depends on how accurately the inferred prototype
assignments reflect target-domain semantics.
Imperfect alignment may introduce conditioning noise and
compromise generation fidelity.
ORBIT instead explicitly organizes temporal representations
and their generation order by frequency, providing a structured
inductive bias for modeling heterogeneous temporal patterns.
In summary, frequency-hierarchical representation learning
and band-level parallel generation contribute to ORBIT's
improved generation quality and fidelity.

\noindent\textbf{Cross-Domain Generation.}
Benefiting from classifier-free guidance (CFG), HALO supports unconditional generation and can be used to unseen domains
excluded from the multi-domain training set.
Following~\citep{timedp}, we use 3, 10, and 100
samples as both prompts and fine-tuning data for other models, with the MMD and K-L divergence results for a
generation length of 168 reported in Figure~\ref{fig:4}.
ORBIT demonstrates superior and robust cross-domain generation capability, substantially outperforming other unconditional baselines and even methods specifically fine-tuned on the target domains.
Notably, ORBIT also maintains a clear advantage over TimeDP~\cite{timedp}, which leverages samples from unseen target domains as prompts.
This further demonstrates ORBIT’s strong cross-domain generalization capability and highlights its potential as a scalable foundation for large-scale time series generative models.

\subsection{Model Analysis}


\noindent\textbf{Ablation Studies.}
To assess each component, we evaluate three ablations:
\emph{\textbf{Variant 1}} uses a single quantization level;
\emph{\textbf{Variant 2}} quantizes frequency bands independently
without residual error compensation; and
\emph{\textbf{Variant 3}} allocates equal token budgets across bands.
As shown in Table~\ref{tab:2}, ORBIT outperforms all variants
across both sequence lengths and metrics.
The degradation of Variants 1 and 2 supports the importance
of hierarchical representations and progressive error correction.
Variant 3 incurs the largest performance drop, suggesting
that frequency-dependent token allocation is essential for
balancing the representation of global structures and
fine-grained details.
Overall, these results validate the effectiveness of
ORBIT's frequency-hierarchical quantization design.
\begin{table}[t]
    \centering
    \setlength{\tabcolsep}{3pt}
    \renewcommand{\arraystretch}{0.7}
    \caption{Ablation study results.
    We report MMD and K-L averaged across datasets for
    long-sequence time series generation task.
    Improvement denotes the relative reduction achieved by the
    full model over each variant, computed from the average
    scores across lengths 96 and 168.
    \label{tab:2}}
    \begin{tabular}{c|c|c|c|c|c}
    \toprule[1.5pt]
    Metrics &
    Length &
    $\mathrm{ORBIT}_{\mathrm{Full}}$ &
    Variant 1 &
    Variant 2 &
    Variant 3 \\
    \midrule[0.5pt]

    \multirow{2}{*}{MMD}
    & 96
    & \textbf{0.059}
    & 0.074
    & 0.079
    & 0.107 \\

    & 168
    & \textbf{0.068}
    & 0.108
    & 0.095
    & 0.124 \\

    \midrule[0.5pt]
    
\multicolumn{2}{c|}{Improvement}
& --
& 30.22\%
& 27.01\%
& 45.02\% \\

    \midrule[0.5pt]

    \multirow{2}{*}{K-L}
    & 96
    & \textbf{0.271}
    & 0.327
    & 0.336
    & 0.354 \\

    & 168
    & \textbf{0.202}
    & 0.228
    & 0.243
    & 0.302 \\

    \midrule[0.5pt]
    
\multicolumn{2}{c|}{Improvement}
& --
& 14.77\%
& 18.31\%
& 27.90\% \\

    \bottomrule[1.5pt]
    \end{tabular}
\end{table}

\begin{figure}
    \centering
    \includegraphics[width=1.\linewidth]{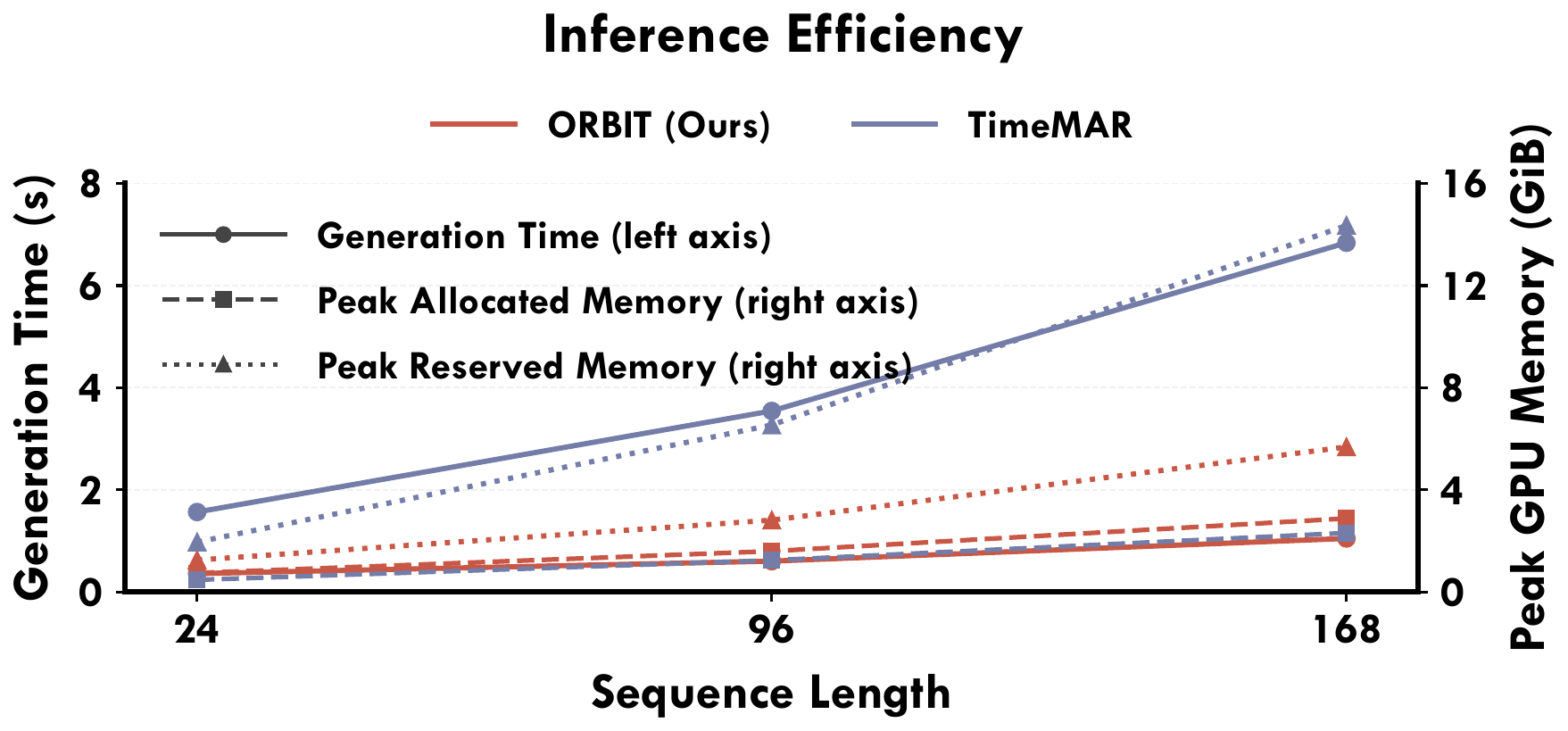}
    \vspace{-14pt}
    \caption{Comparison of generation time, peak allocated memory, and
peak reserved memory between ORBIT and TimeMAR~\citep{timemar} for generating
5,000 samples on the Electricity dataset with a fixed batch
size of 1,000.}
    \label{fig:5}
\end{figure}

\noindent\textbf{Inference Efficiency Analysis.}
To investigate the inference efficiency advantages of ORBIT, we conduct a comparative efficiency analysis.
Specifically, we benchmark ORBIT against TimeMAR~\citep{timemar}, a representative state-of-the-art baseline.
As shown in Figure~\ref{fig:5}, ORBIT consistently achieves superior efficiency across different sequence lengths, with lower generation time, peak allocated memory, and peak reserved memory.
This advantage is primarily attributed to ORBIT's ability to perform parallel decoding at the frequency-band level.
The efficiency advantages of ORBIT revealed by the efficiency evaluation strongly underscore the necessity of further exploring more accurate and efficient parallel decoding strategies.

\noindent\textbf{Visualization.}
To qualitatively assess generation quality, we use t-SNE~\citep{tsne}
to project real and synthetic time series
from representative datasets into a two-dimensional space,
visualizing their distributional overlap.
As shown in Figure~6, ORBIT exhibits greater overlap between
real and synthetic samples in the t-SNE space across
representative datasets, suggesting closer alignment
of their underlying representations.
\begin{figure}
    \centering
    \includegraphics[width=0.48\linewidth]{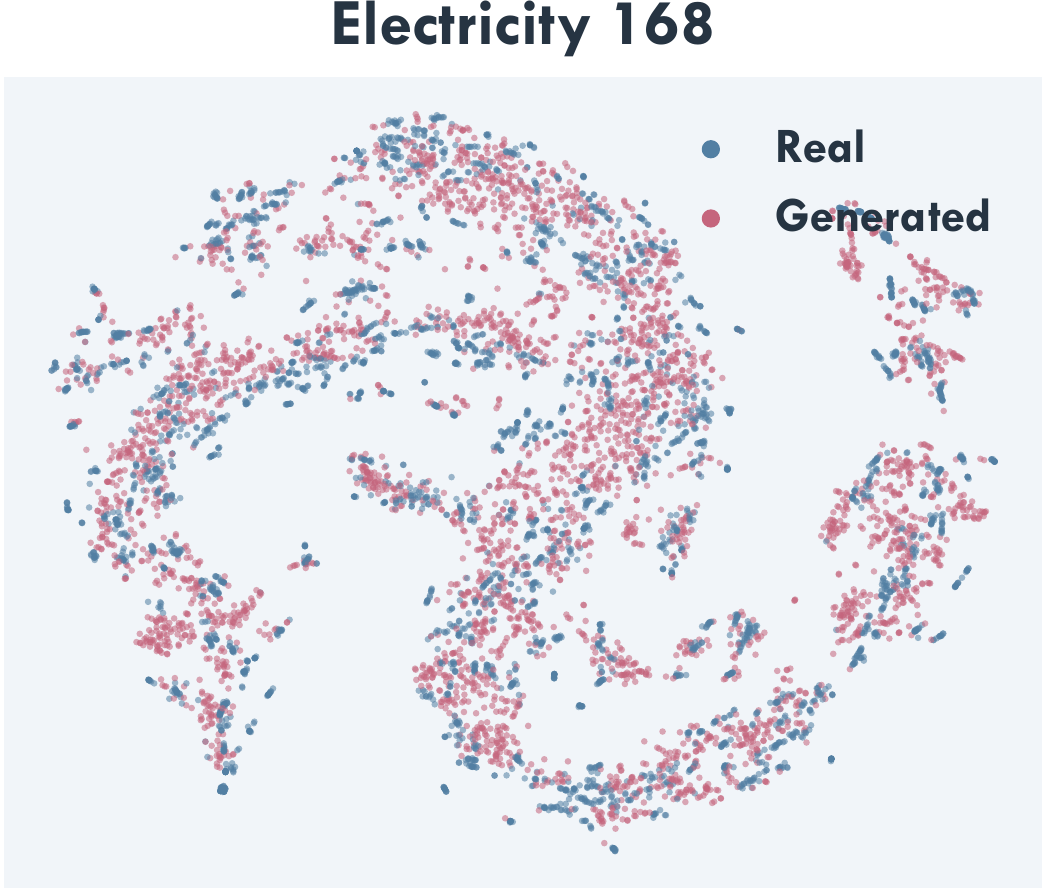}
    \includegraphics[width=0.48\linewidth]{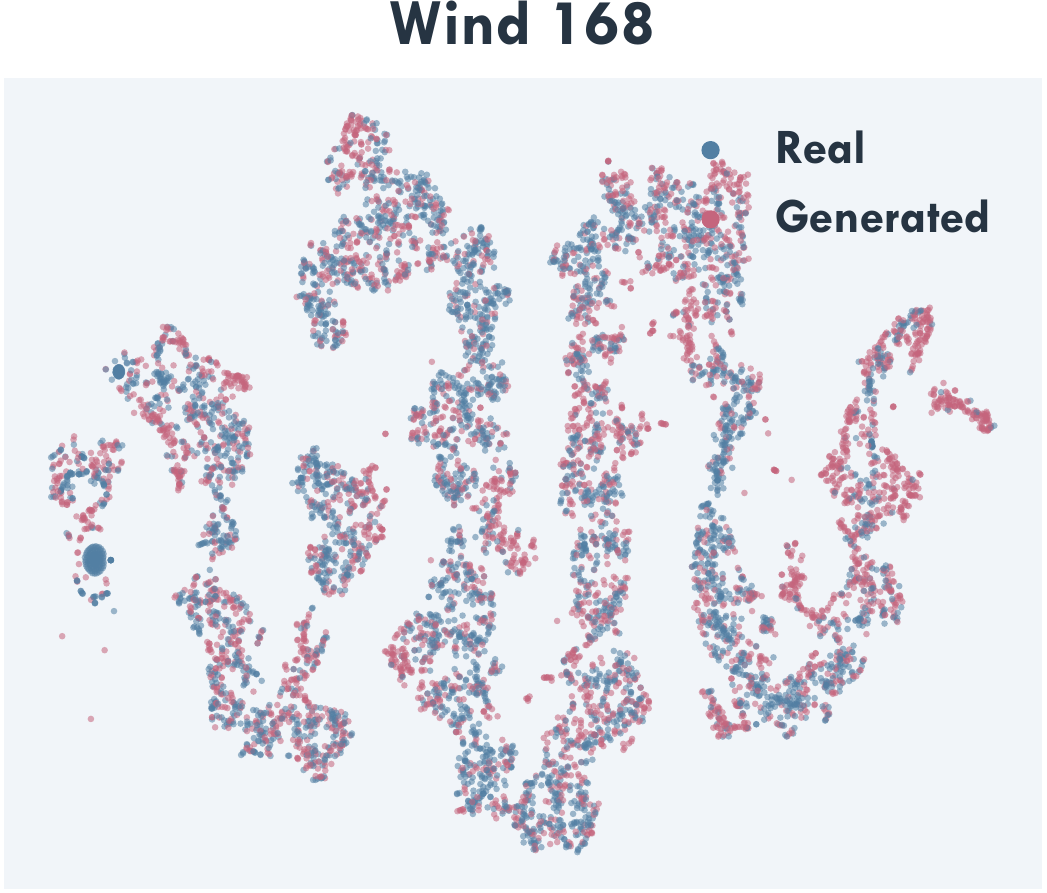}
    \caption{t-SNE visualization of real and synthetic time series samples on the Electricity and Wind benchmarks with a sequence length of 168.
    Additional qualitative visualizations are provided in Appendix~\ref{app:d}.
    \label{fig:6}}
    
\end{figure}

\section{Conclusions}
In this paper, we propose ORBIT, a unified deep generative framework, which enables high-quality and high-fidelity time series generation.
Its Frequency-Hierarchical VQ-VAE combines frequency-band
decomposition with residual quantization to better learn both
global temporal structures and fine-grained temporal details.
Its Frequency-Band Autoregressive Transformer models
cross-band correlations through next frequency-band prediction,
enabling low-to-high-frequency generation with parallel
decoding within each band.
Experiments across 12 datasets demonstrate state-of-the-art
overall generation performance and improved inference
efficiency, while ablation studies support the effectiveness
of the core components.
These findings highlight the potential of frequency-hierarchical
representations and effective autoregressive modeling for
high-quality and efficient time series generation.

\bibliographystyle{ACM-Reference-Format}
\bibliography{sample-base}

\appendix
\clearpage
\section{Dataset Description}\label{app:a}

We provide detailed descriptions of the 12 datasets used
for model training and in-domain generation evaluation.

    \noindent\textbf{Electricity}.
    This dataset contains hourly electricity consumption
    measurements, expressed in kilowatts (kW), from 321 clients
    between 2012 and 2014. It is sourced from the UCI Machine
    Learning Repository.

    \noindent\textbf{Solar}.
    This dataset contains 137 time series of hourly solar
    power production in Alabama, USA, during 2006.

    \noindent\textbf{Wind}.
    This dataset comprises a single long time series of wind
    power production, measured in megawatts (MW) at 4-second
    intervals starting on August 1, 2019. It is sourced from
    the Australian Energy Market Operator (AEMO) online platform.

    \noindent\textbf{Traffic}.
    This dataset contains 440 daily records spanning 15 months.
    It records freeway lane occupancy rates in the San Francisco
    Bay Area, with values ranging from 0 to 1.

    \noindent\textbf{Taxi}.
    This dataset contains spatiotemporal records of New York
    City taxi rides across 1,214 locations, collected at
    30-minute intervals during January 2015 and January 2016.

    \noindent\textbf{Pedestrian}.
    This dataset contains hourly pedestrian counts from
    66 sensors across Melbourne, Australia, starting in
    May 2009. Although the source dataset is continuously
    updated, we use the version containing observations
    through April 30, 2020.

    \noindent\textbf{Air Quality}.
    Originally used in the KDD Cup 2018 forecasting competition,
    this dataset contains hourly air-quality measurements
    from 59 monitoring stations: 35 in Beijing and 24 in London.
    The observations span January 1, 2017 to March 31, 2018,
    and cover PM$_{2.5}$, PM$_{10}$, NO$_2$, CO, O$_3$, and SO$_2$,
    yielding 270 time series indexed by city, station, and
    pollutant. Leading missing values are filled with zeros,
    and subsequent missing values are imputed using the last
    observation carried forward (LOCF) strategy.

    \noindent\textbf{Temperature}.
    This dataset contains 32,072 daily records of temperature
    observations and rainfall forecasts collected by the
    Australian Bureau of Meteorology from 422 weather stations
    across Australia between May 2, 2015 and April 26, 2017.
    Missing values are replaced with zeros.
    We use the mean temperature variable in our experiments.

    \noindent\textbf{Rain}.
    This dataset shares the same source as Temperature.
    We use the rainfall variable to construct the time series.

    \noindent\textbf{NN5}.
    Originally used in the NN5 forecasting competition,
    this dataset contains 111 time series of daily cash
    withdrawals from automated teller machines (ATMs)
    in the United Kingdom. Missing values are imputed
    using the median of observations falling on the same
    day of the week within each series.

    \noindent\textbf{Fred-MD}.
    This dataset contains 107 monthly macroeconomic time
    series obtained from the Federal Reserve's FRED-MD database.
    Following common practice, we apply differencing and
    logarithmic transformations as appropriate for each series.

    \noindent\textbf{Exchange}.
    This dataset contains daily exchange-rate time series
    for eight currencies.

We additionally evaluate unseen-domain generation on the
following two datasets, which are excluded from model training.

    \noindent\textbf{Stock}.
    This dataset contains daily stock prices for Google
    (GOOG), listed on the NASDAQ stock exchange.

    \noindent\textbf{Web}.
    Originally used in the Kaggle Wikipedia Web Traffic
    Forecasting competition, this dataset contains 145,063
    daily time series of Wikipedia page views spanning
    July 1, 2015 to September 10, 2017.
    Missing values are replaced with zeros.

\section{Baseline Details}\label{app:b}

We compare HALO with the following time series generation baselines,
covering adversarial, variational, diffusion-based, and discrete-token
generative approaches.
Table~\ref{tab:3} summarizes the code repositories of the baseline methods.
\begin{table}[h]
    \centering
    \setlength{\tabcolsep}{0.5pt}
    \renewcommand{\arraystretch}{0.4}
    \caption{\small Code repositories of the baseline methods.
    TimeVQVAE-C shares the implementation of TimeVQVAE.}
    \label{tab:3}
    \small
    \resizebox{0.45\textwidth}{!}{%
    \begin{tabular}{lc}
        \toprule[1.5pt]
        \textbf{Baseline} & \textbf{Code Repository} \\
        \midrule[0.5pt]
        TimeMAR
        & \url{https://github.com/decisionintelligence/TimeMAR} \\
        \midrule[0.5pt]
        TimeDP
        & \url{https://github.com/microsoft/TimeCraft/tree/main/TimeDP} \\
        \midrule[0.5pt]
        TimeGAN
        & \url{https://github.com/jsyoon0823/TimeGAN} \\
        \midrule[0.5pt]
        GT-GAN
        & \url{https://github.com/Jinsung-Jeon/GT-GAN} \\
        \midrule[0.5pt]
        TimeVAE
        & \url{https://github.com/abudesai/timeVAE} \\
        \midrule[0.5pt]
        TimeVQVAE
        & \url{https://github.com/ML4ITS/TimeVQVAE} \\
        \midrule[0.5pt]
        TimeVQVAE-C
        & \url{https://github.com/ML4ITS/TimeVQVAE} \\
        
        \bottomrule[1.5pt]
    \end{tabular}}
\end{table}

\paragraph{\textbf{TimeMAR}~\citep{timemar}.}
TimeMAR is a two-stage generative framework that combines multi-scale
discrete representations with autoregressive modeling.
It employs a dual-path VQ-VAE to encode decomposed trend and seasonal
components into discrete tokens at multiple temporal resolutions,
followed by a Transformer that generates these tokens in a
coarse-to-fine manner.
Coarse seasonal representations further guide the reconstruction
of fine-grained seasonal patterns.
Following the experimental protocol of~\citep{timedp}, we adapt it to use domain labels as prompts.

\paragraph{\textbf{TimeDP}~\citep{timedp}.}
TimeDP is a diffusion-based framework for multi-domain time series
generation.
It learns a shared set of semantic prototypes to represent temporal
patterns and employs a prototype assignment module to construct
domain prompts from input time series.
These prompts condition the diffusion model, enabling domain-specific
generation using a small set of examples from the target domain.

\paragraph{\textbf{TimeGAN}~\citep{TIMEGAN}.}
TimeGAN combines adversarial learning with a supervised temporal
objective in a learned embedding space.
Its embedding and recovery networks provide latent representations
for generation, while the supervised objective encourages the model
to preserve temporal dependencies between consecutive latent states.

\paragraph{\textbf{GT-GAN}~\citep{gtgan}.}
GT-GAN is an adversarial framework designed to synthesize both
regularly and irregularly sampled time series.
It integrates neural ordinary and controlled differential equations
with continuous-time flow processes to capture temporal dynamics
under different sampling patterns.

\paragraph{\textbf{TimeVAE}~\citep{timevae}.}
TimeVAE is a variational autoencoder for multivariate time series
generation.
It combines continuous latent representations with an interpretable
decoder that explicitly models level, trend, and seasonal components,
allowing generated sequences to reflect these temporal structures.

\paragraph{\textbf{TimeVQVAE}~\citep{timevqvae}.}
TimeVQVAE adopts a two-stage framework based on vector quantization
and bidirectional prior modeling.
It separates time-frequency representations into low- and
high-frequency components and encodes them into discrete tokens.
Bidirectional Transformer priors learn the token distributions
through masked modeling, enabling generation via iterative
masked decoding.

\paragraph{\textbf{TimeVQVAE-C}~\citep{timevqvae,timedp}.}
TimeVQVAE-C denotes the class-conditional variant of TimeVQVAE.
It retains the time-frequency tokenization framework while
conditioning the bidirectional token prior on class labels,
enabling the generation of time series associated with a
specified class.

\section{More Implementation Details}\label{app:c}
We provide detailed hyperparameter settings for ORBIT.
For the first-stage FH-VQ-VAE, we report the number and dimensionality of encoder-produced continuous latent representations, the frequency-band partitioning scheme, the token budget for each frequency band, and the codebook size.
Table~\ref{tab:4} presents the detailed hyperparameter settings.
For the second-stage FBAR, we report its backbone width, number of Transformer layers, number of attention heads, and hidden dimensionality of the feed-forward network (FFN). 
Detailed hyperparameter settings are presented in Table~\ref{tab:5}.
Both stages are optimized using the AdamW optimizer~\citep{adam} with batch size 256.

\begin{table}[h]
    \centering
    \caption{Detailed hyperparameter settings of FH-VQ-VAE.
    \(M\), \(d\), \(B\), \(n_b\), and \(K\) denote the number of continuous latents produced by the encoder, the hidden dimensionality, the number of frequency bands, the number of quantization tokens allocated to the \(b\)-th frequency band, and the total number of codewords in the codebook, respectively.
    \label{tab:4}}
    \begin{tabular}{c|c|c|c|c|c|c}
    \toprule[1.5PT]
     \multicolumn{2}{c|}{Hyperparameter} &
     $M$ &
     $d$ &
     $B$ &
     $n_{b}$ &
     $K$ \\
     \midrule[0.5pt]
     \multirow{3}{*}{Length} &
     24 &
     6 &
     16  &
     6 &
     [1, 2, 3, 4, 5, 6] &
     2048  \\
     
    \cmidrule{2-7}
     
     &
     96 &
     24 &
     16  &
     6 &
     [1, 2, 3, 6, 12, 24] &
     2048  \\

     \cmidrule{2-7}

     &
     168 &
     42 &
     16  &
     6 &
     [1, 2, 6, 12, 24, 42] &
     2048  \\

     \bottomrule[1.5pt]
     
    \end{tabular}
    
\end{table}

\begin{table}[h]
    \centering
    \caption{Detailed hyperparameter settings of FBAR.
    \(D\), \(Layer\), \(Head\), and \(FNN\) denote the backbone width, number of Transformer layers, number of attention heads, and hidden dimensionality of the feed-forward network, respectively.
    \label{tab:5}}

    \begin{tabular}{c|c|c|c|c|c}
    \toprule[1.5PT]
     \multicolumn{2}{c|}{Hyperparameter} &
     $D$ &
     $Layer$ &
     $Head$ &
     $FNN$ \\
     \midrule[0.5pt]
     \multirow{3}{*}{Length} &
     24 &
     512 &
     4  &
     8 &
     4*$D$   \\
     
    \cmidrule{2-6}
     
     &
     96 &
     512 &
     4  &
     8 &
     4*$D$   \\

     \cmidrule{2-6}

     &
     168 &
     512 &
     4  &
     8 &
     4*$D$   \\

     \bottomrule[1.5pt]
     
    \end{tabular}
    
\end{table}

\section{Further Visualizations}\label{app:d}

\begin{figure}[h]
    \centering
    \includegraphics[width=0.48\linewidth]{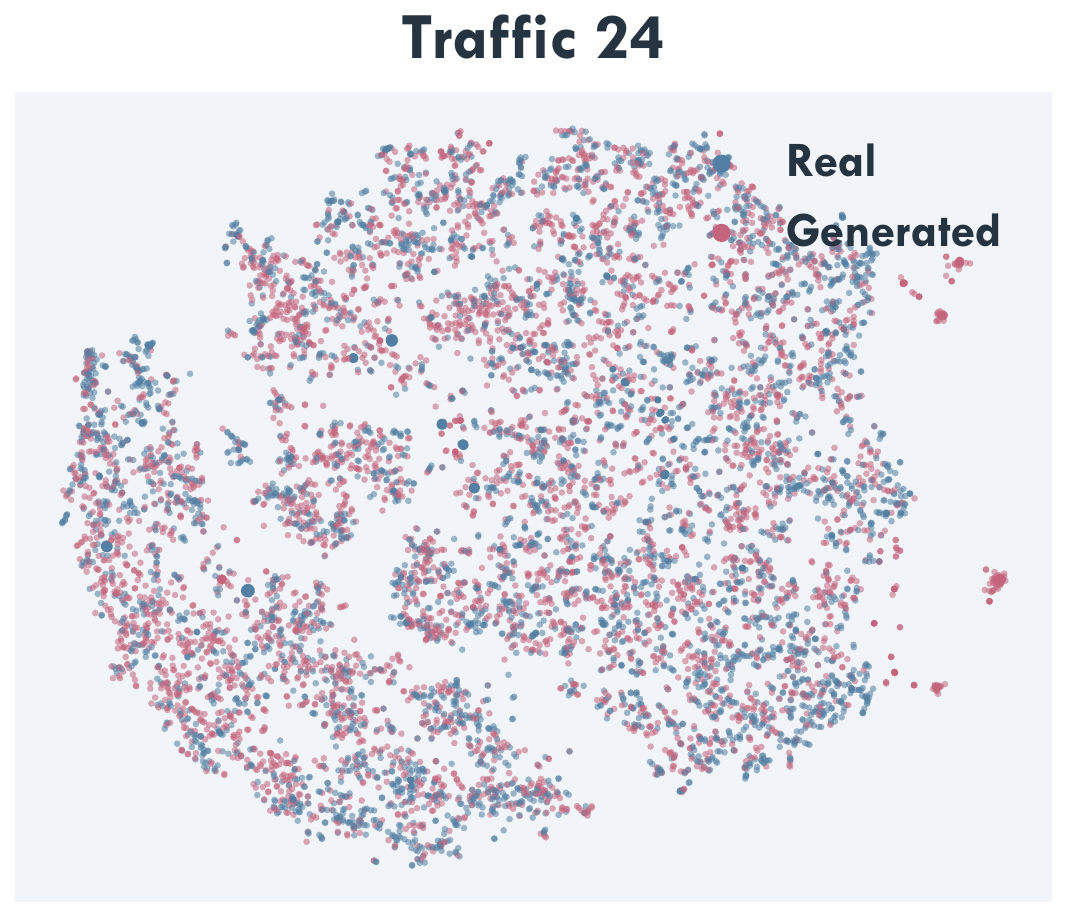}
    \includegraphics[width=0.48\linewidth]{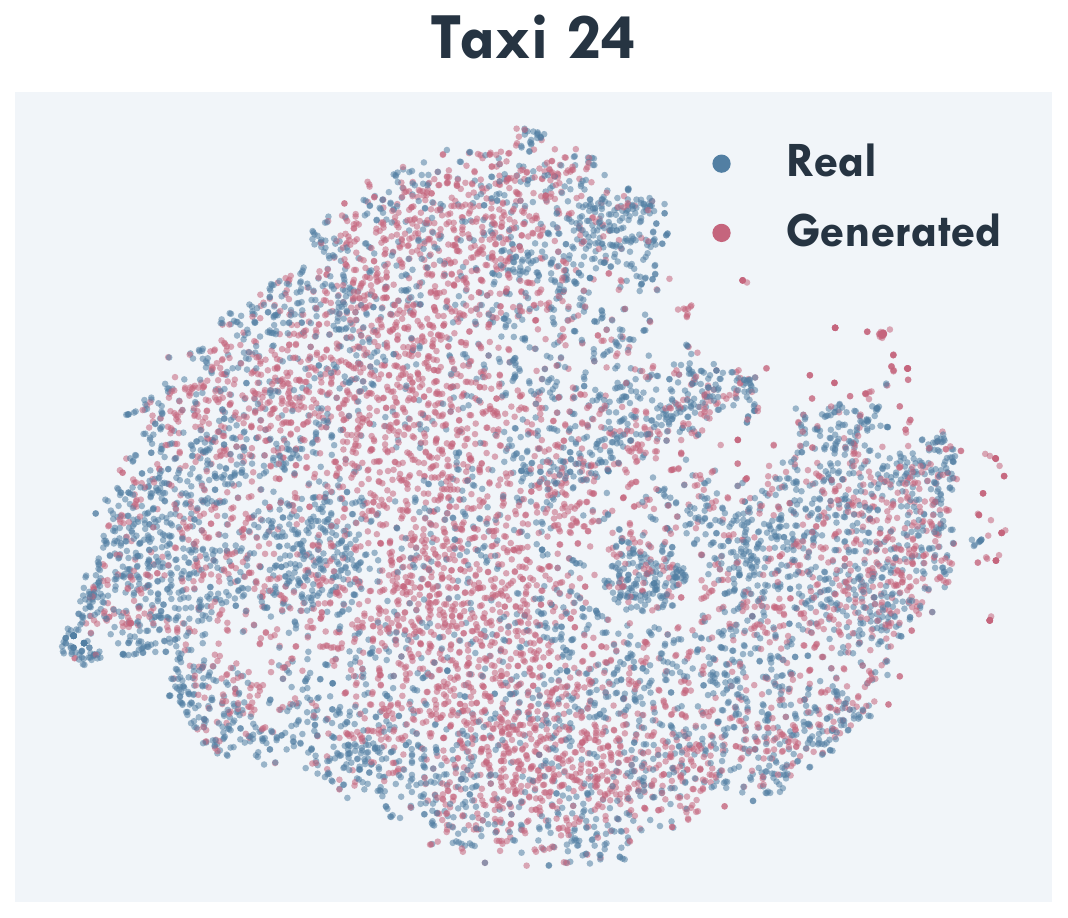} \\
    
    \includegraphics[width=0.48\linewidth]{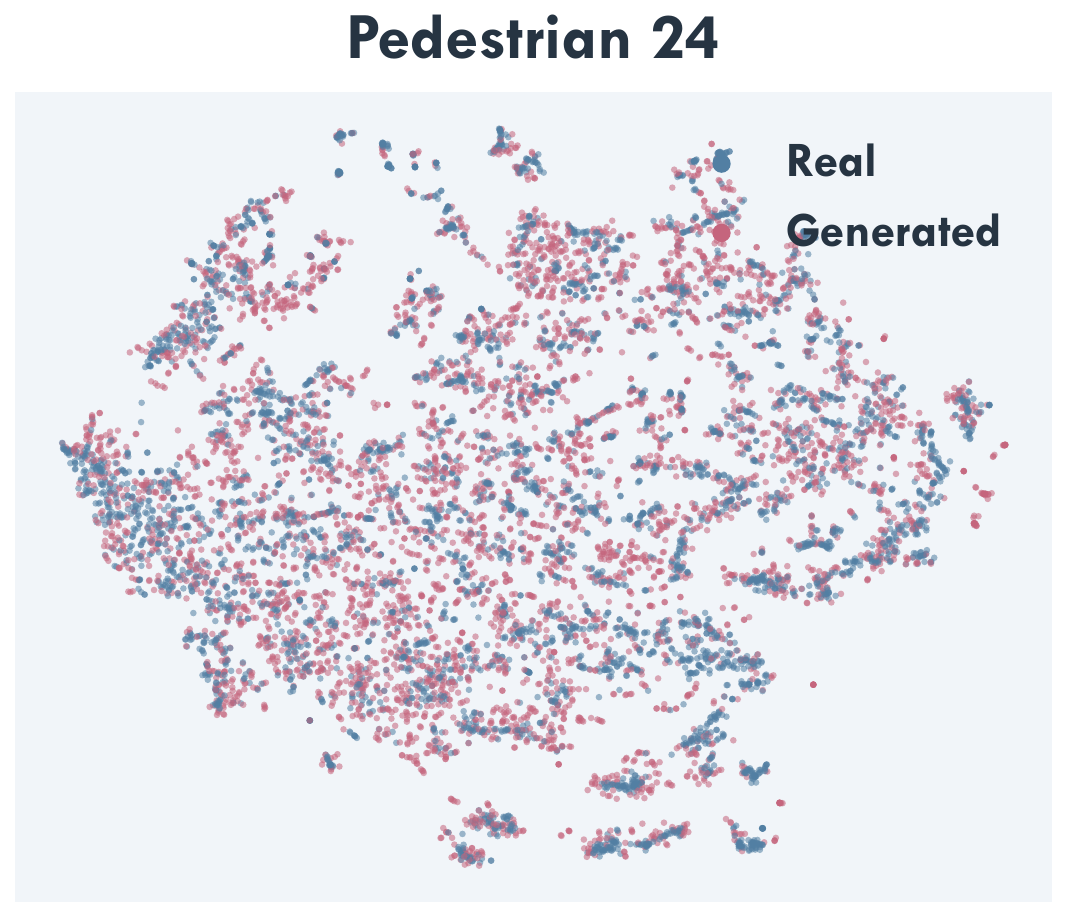}
    \includegraphics[width=0.48\linewidth]{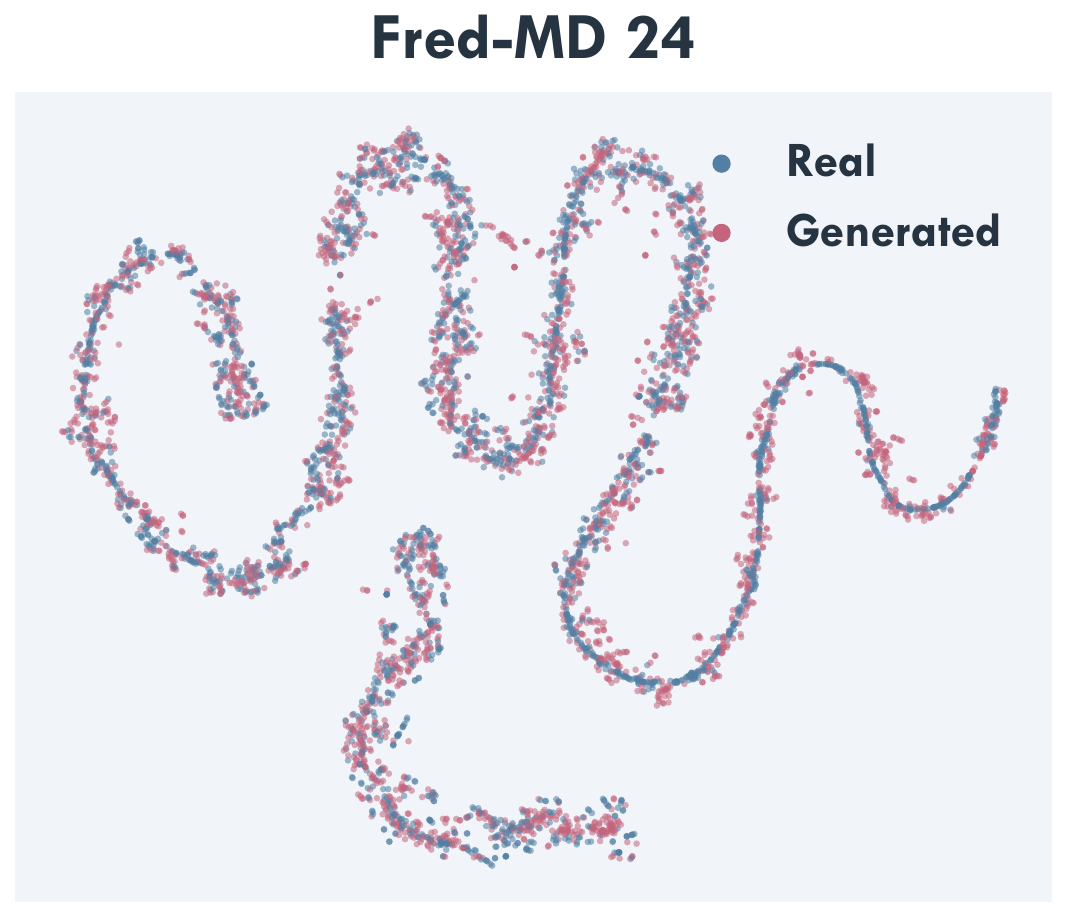}
    \caption{t-SNE visualization of real and synthetic time series samples on the Traffic, Taxi, Pedestrian, and Fred-MD benchmarks with a sequence length of 24.
    \label{fig:7}}
\end{figure}

\begin{figure}[h]
    \centering
    \includegraphics[width=0.48\linewidth]{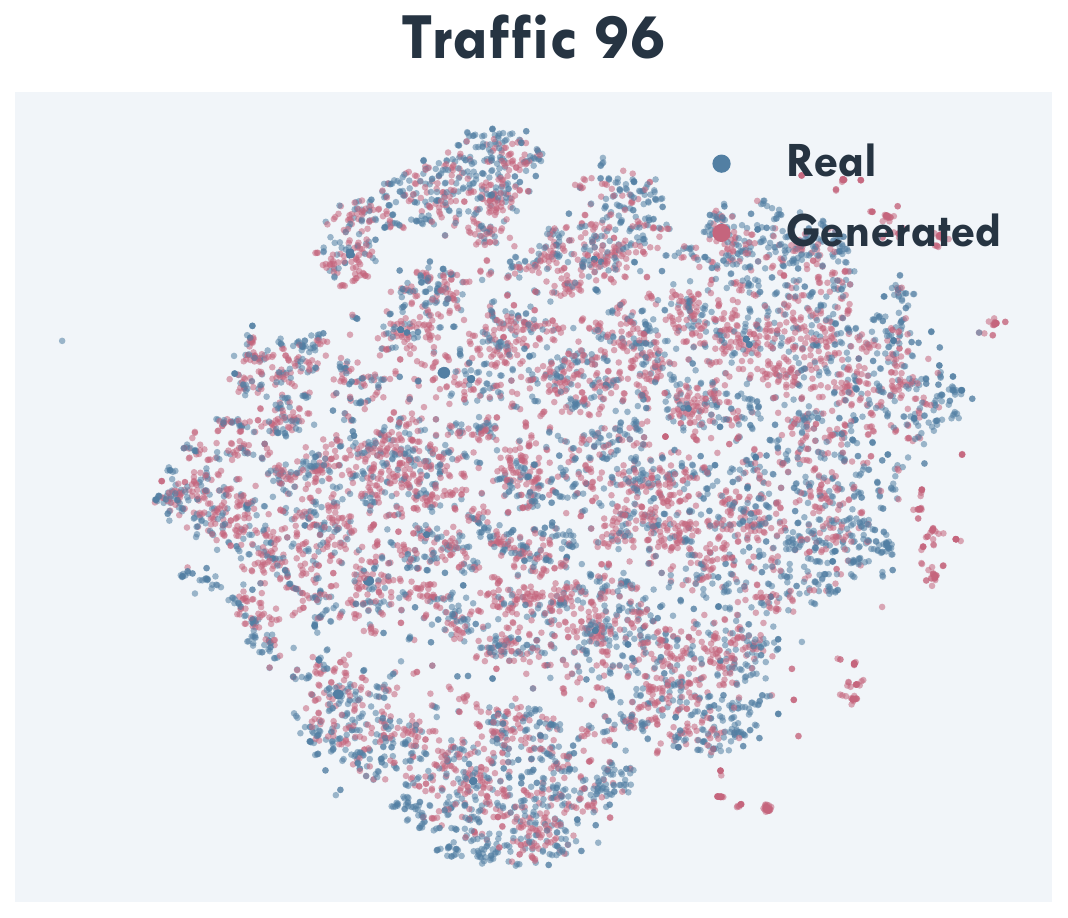}
    \includegraphics[width=0.48\linewidth]{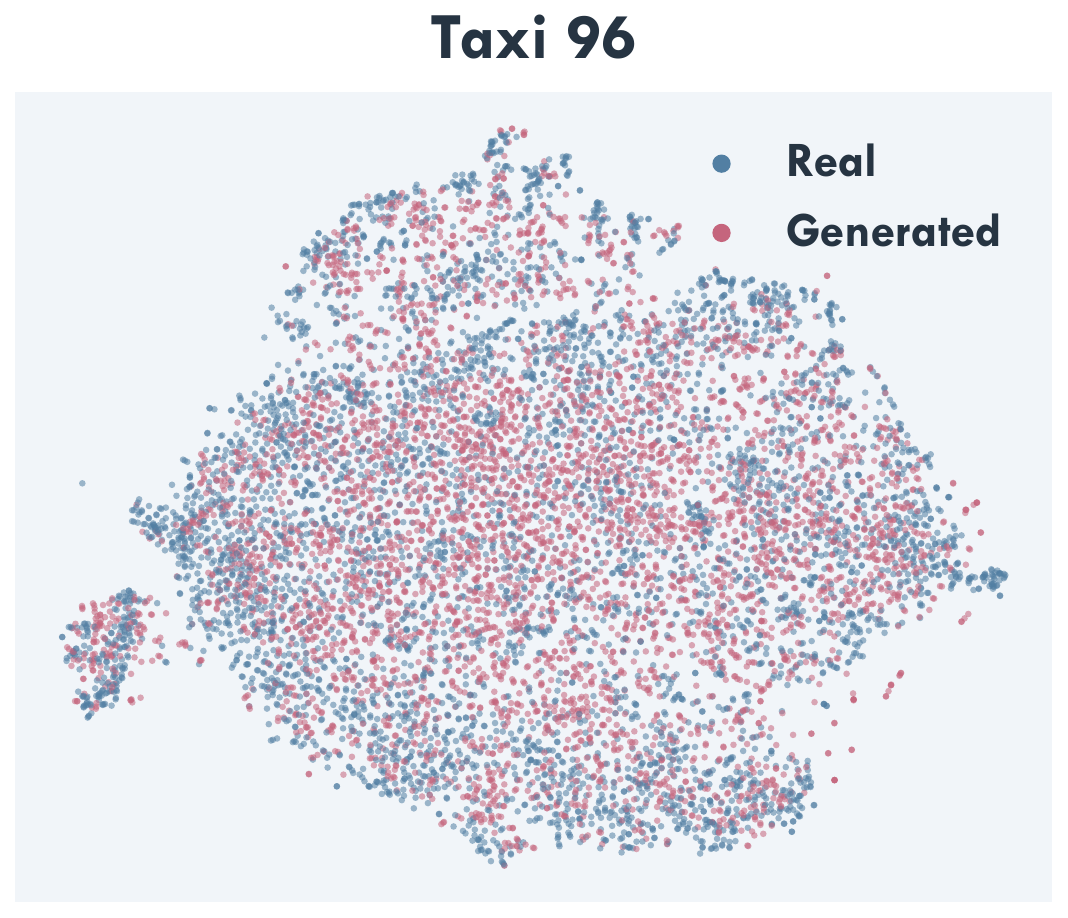} \\
    
    \includegraphics[width=0.48\linewidth]{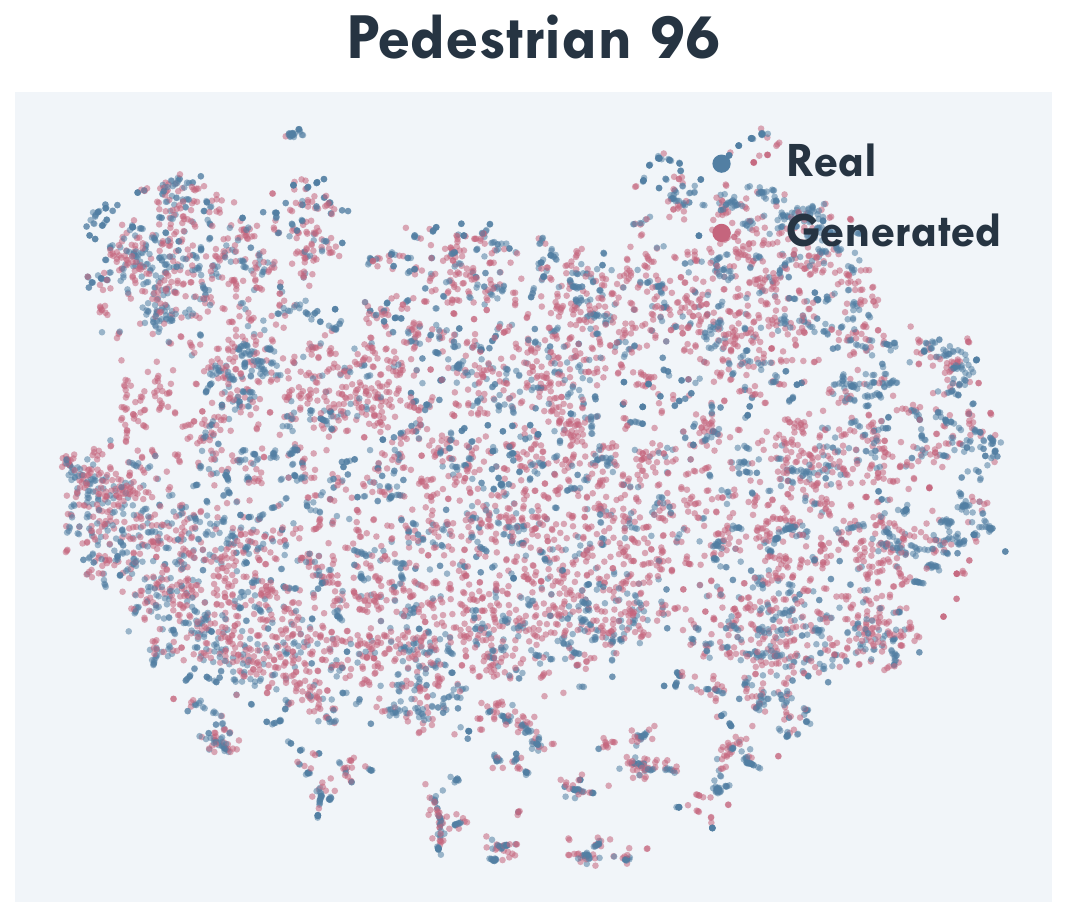}
    \includegraphics[width=0.48\linewidth]{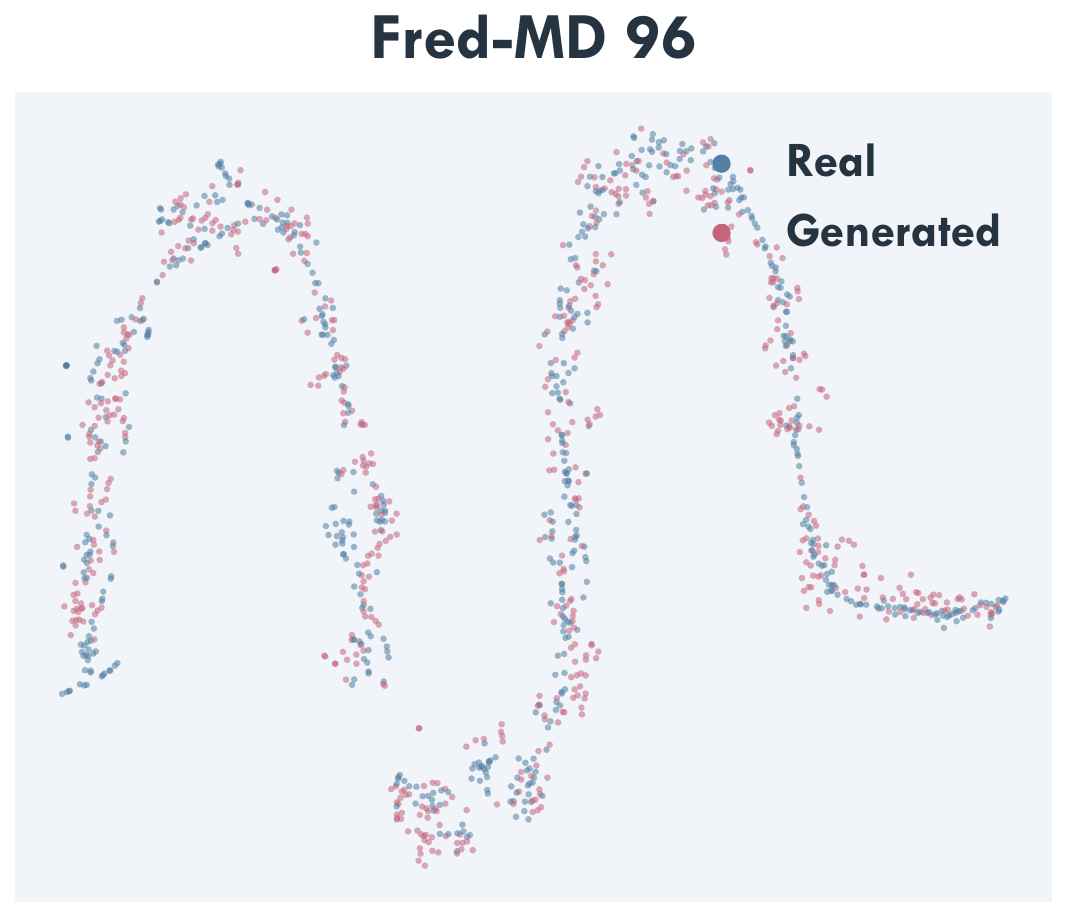}
    \caption{t-SNE visualization of real and synthetic time series samples on the Traffic, Taxi, Pedestrian, and Fred-MD benchmarks with a sequence length of 96.
    \label{fig:8}}
\end{figure}

\begin{figure}[h]
    \centering
    \includegraphics[width=0.48\linewidth]{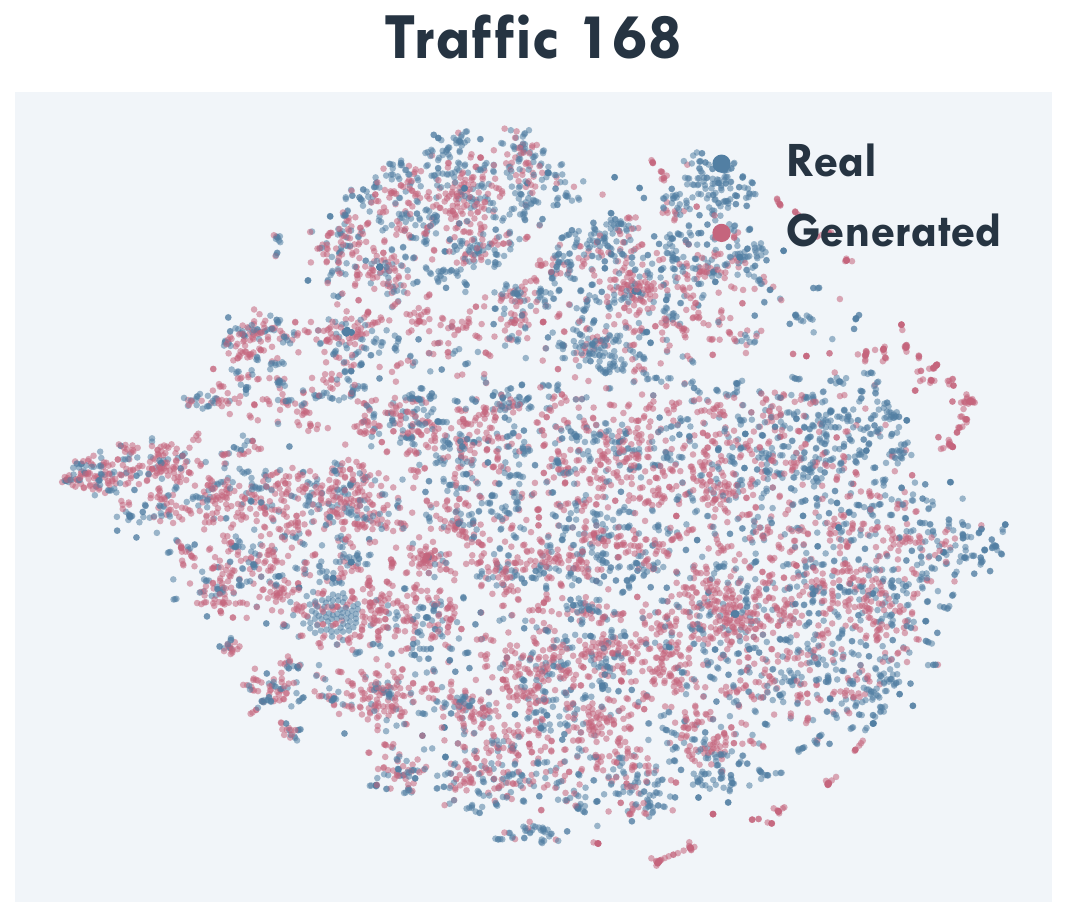}
    \includegraphics[width=0.48\linewidth]{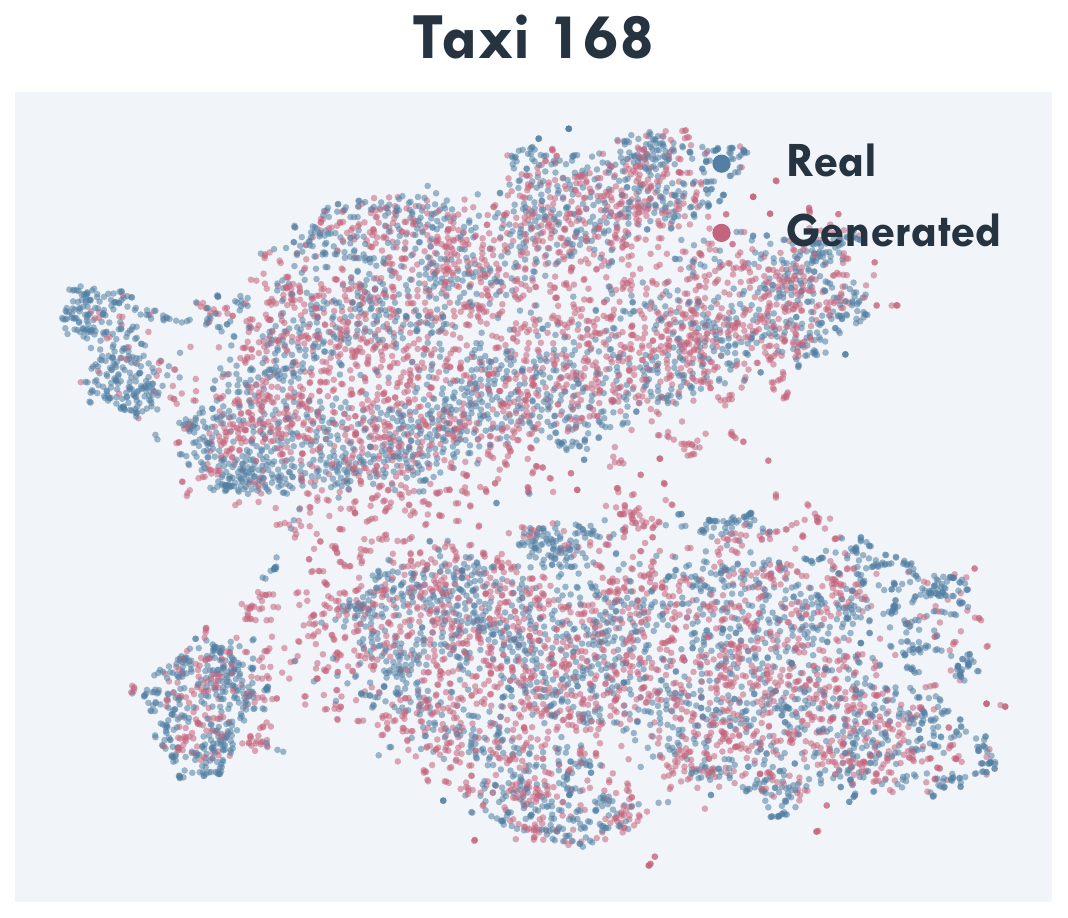} \\
    
    \includegraphics[width=0.48\linewidth]{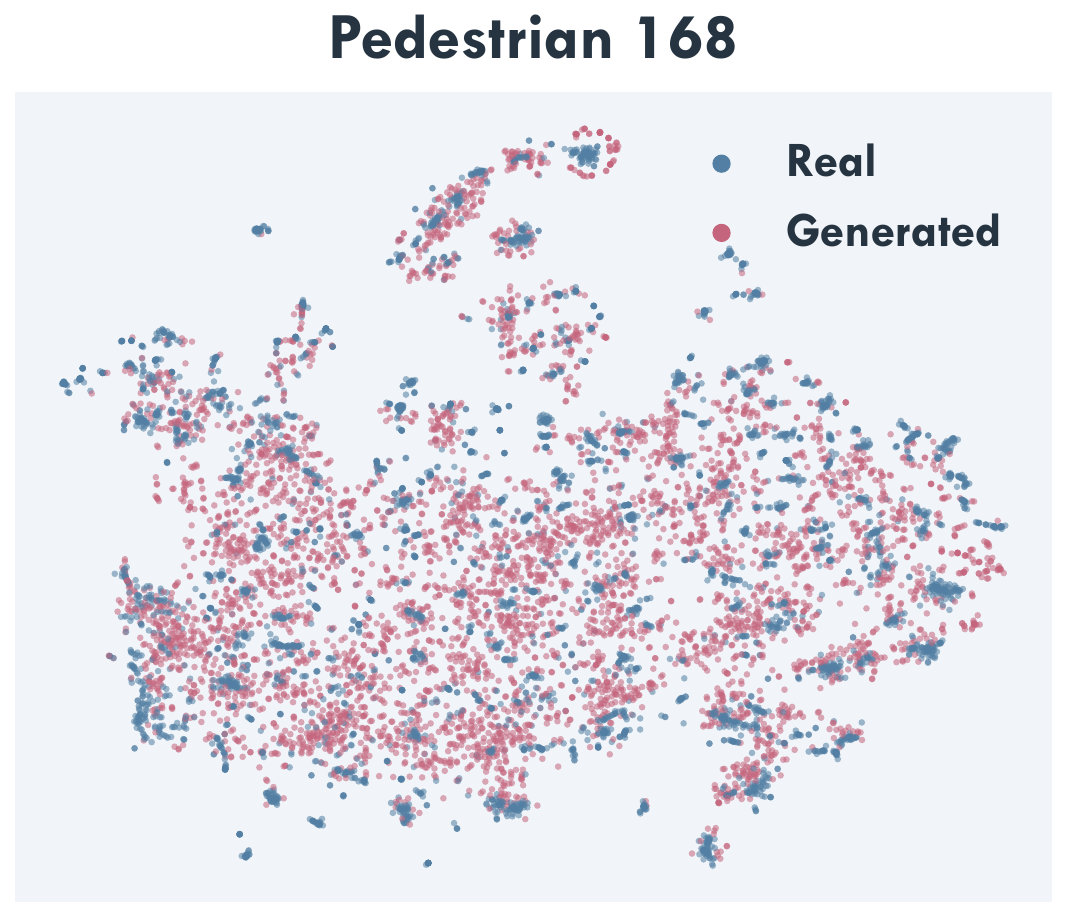}
    \includegraphics[width=0.48\linewidth]{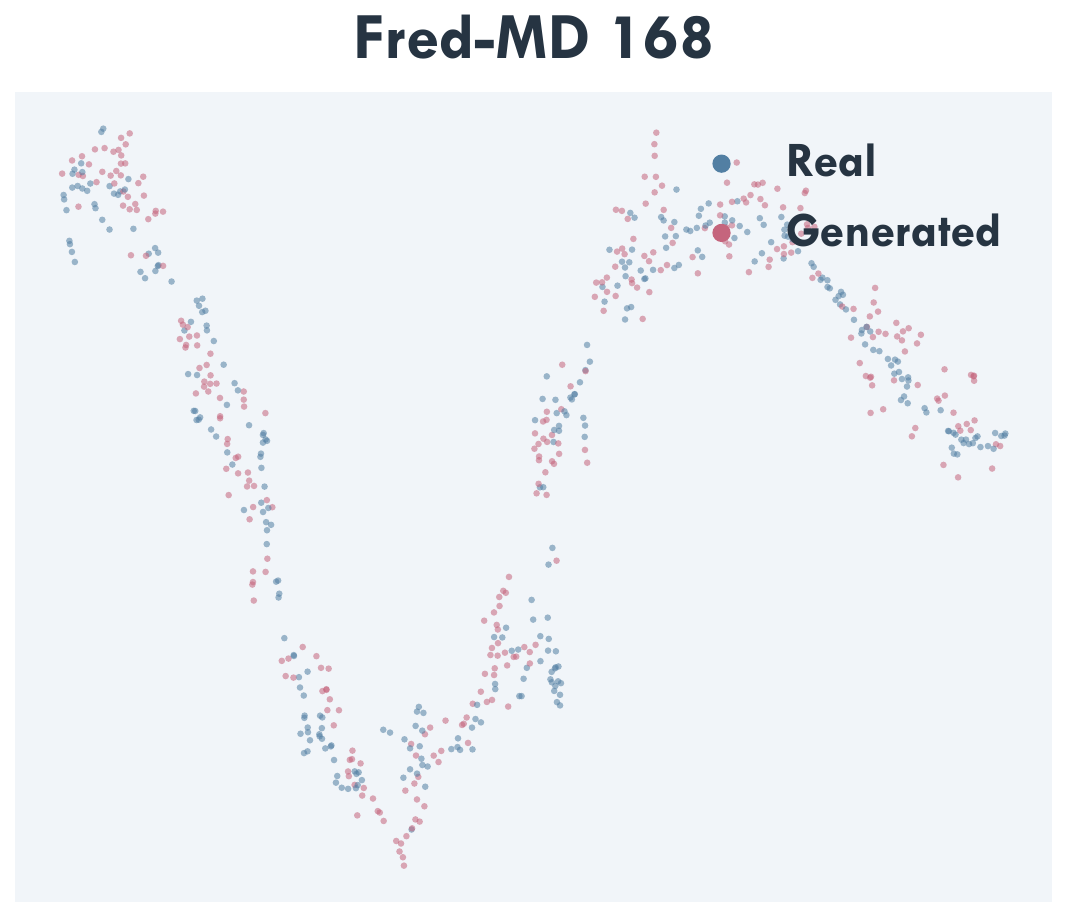}
    \caption{t-SNE visualization of real and synthetic time series samples on the Traffic, Taxi, Pedestrian, and Fred-MD benchmarks with a sequence length of 168.
    \label{fig:9}}
\end{figure}
Figures~\ref{fig:7}-\ref{fig:9} present additional visualizations of generated time series on other benchmarks with sequence lengths of 24, 91, and 168.
As observed, real and synthetic samples exhibit substantial overlap in their primary distribution regions, with no pronounced global shifts. 
Notably, several peripheral clusters, in addition to the central region, contain both real and synthetic samples, suggesting that ORBIT captures diverse sample patterns rather than merely reproducing dominant distribution modes.

\section{Additional Experimental Results}\label{app:e}
In this section, we provide additional experimental results covering sampling temperature sensitivity and a case study on next frequency-band prediction.

\begin{figure}[h]
    \centering
    \includegraphics[width=1.\linewidth]{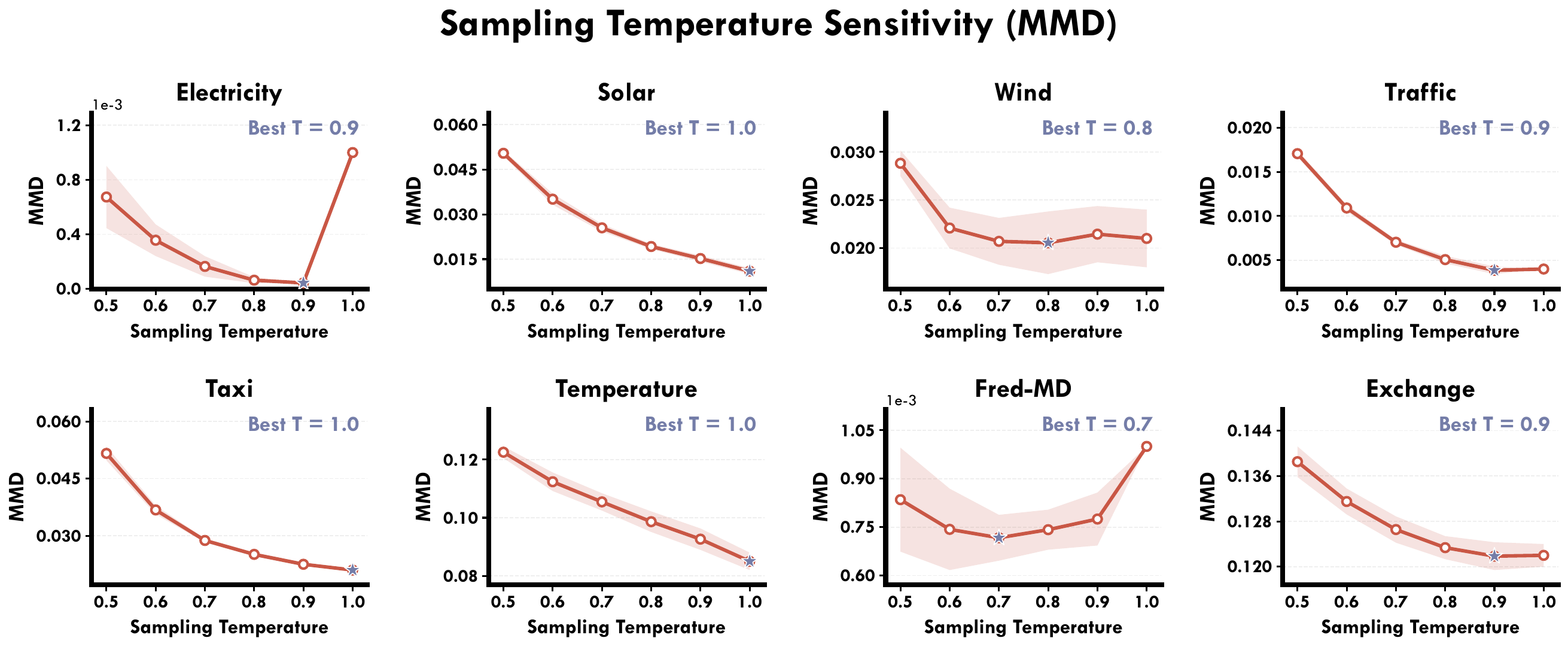}
    \caption{Sensitivity analysis of MMD to sampling temperature across eight benchmark datasets.}
    \label{fig:10}
\end{figure}

\begin{figure}[h]
    \centering
    \includegraphics[width=1.\linewidth]{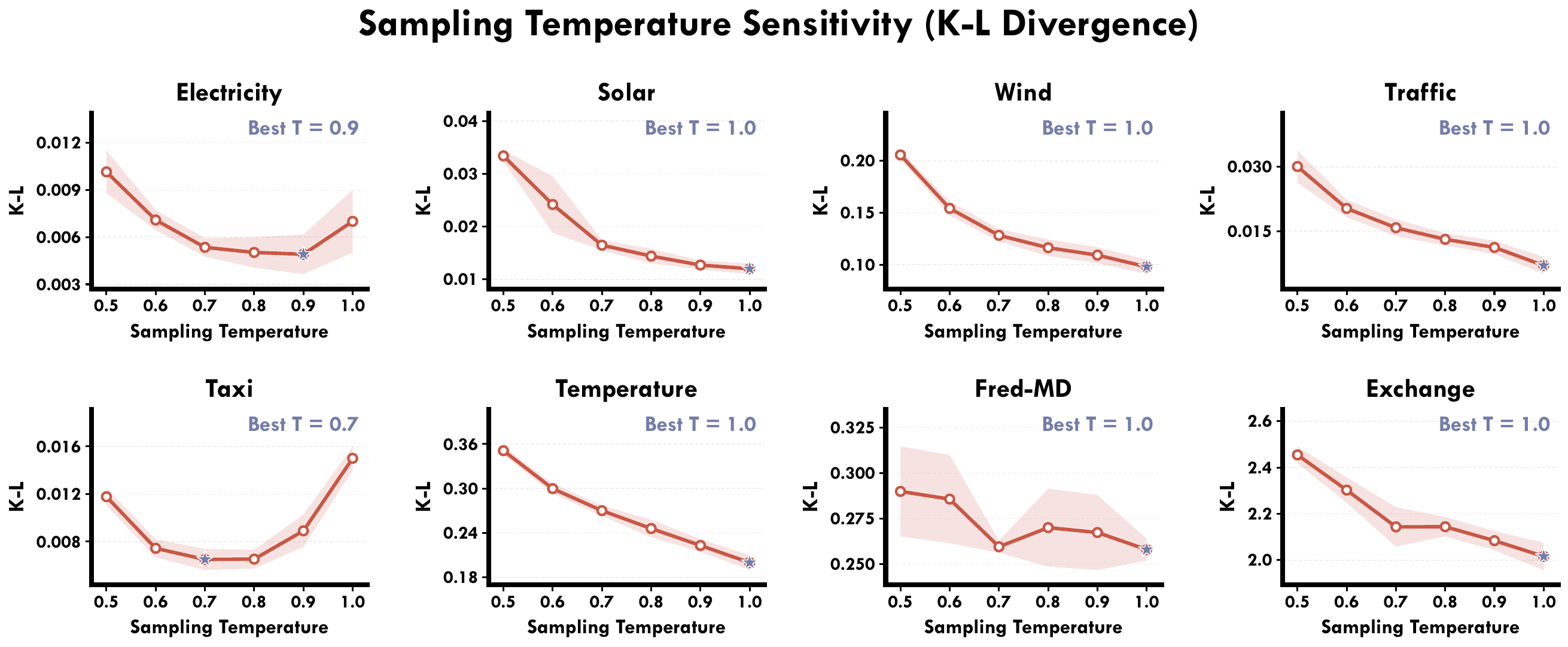}
    \caption{Sensitivity analysis of K-L to sampling temperature across eight benchmark datasets.}
    \label{fig:11}
\end{figure}

\noindent\textbf{Sensitivity to Sampling Temperature.}
As shown in Figures~\ref{fig:10} and~\ref{fig:11}, increasing the temperature generally improves distribution matching, although the optimal setting varies across datasets and metrics.
The lowest mean MMD occurs at temperatures between 0.9 and 1.0 on six of the eight datasets, while the lowest mean K-L occurs in this range on seven datasets.
Solar and Temperature exhibit consistent improvements in both metrics as the temperature increases.
However, Wind and Fred-MD achieve their lowest MMD at temperatures of 0.8 and 0.7, respectively, whereas their K-L is minimized at 1.0.
Taxi exhibits a distinct trade-off: its MMD decreases with increasing temperature, while its K-L reaches its minimum near 0.7 before increasing.
These results support 0.9-1.0 as a favorable operating range for most evaluated datasets, while demonstrating that no single temperature uniformly optimizes both metrics.
Notably, ORBIT achieves state-of-the-art overall generation performance without requiring careful tuning of the sampling temperature, which further highlights the robustness and effectiveness of our framework.

\begin{figure}
    \centering
    \includegraphics[width=1.\linewidth]{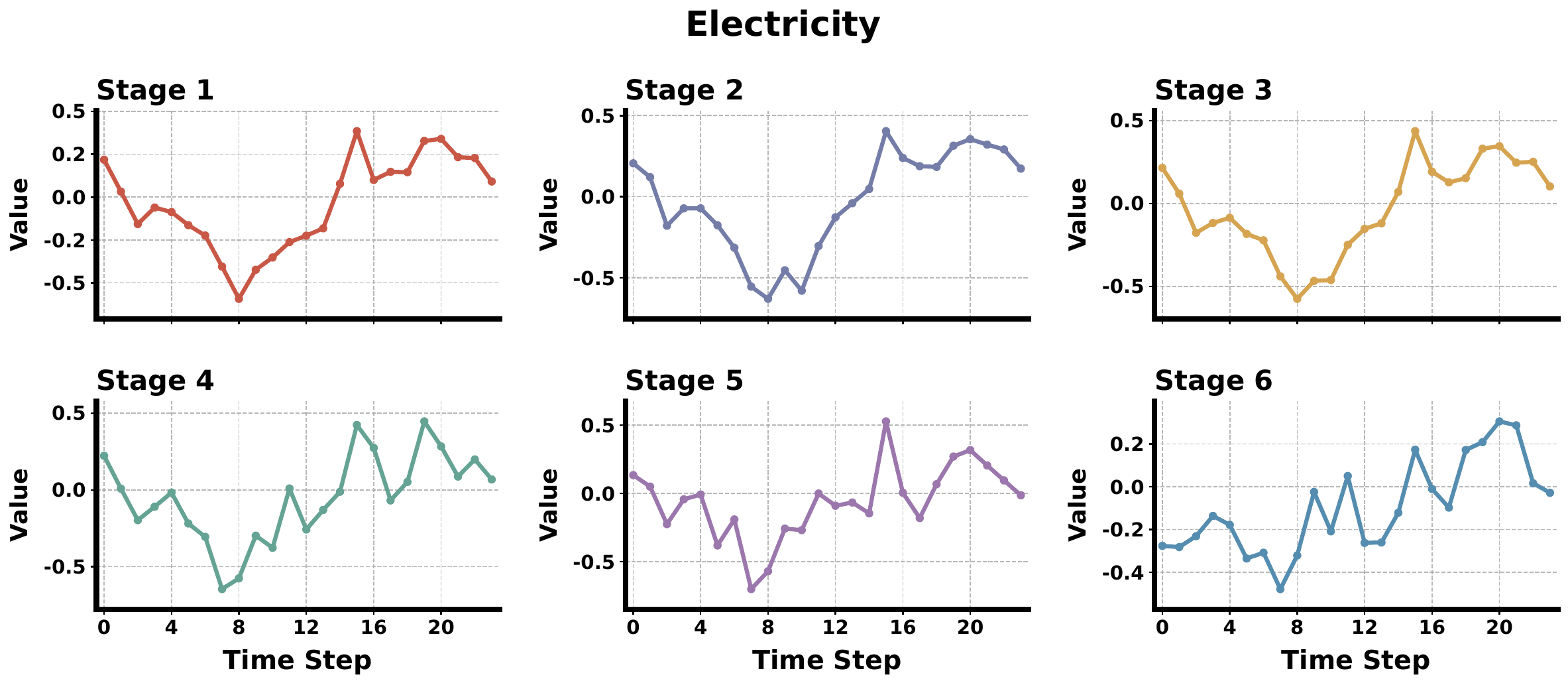} \\
    \vspace{8pt}
    \includegraphics[width=1.\linewidth]{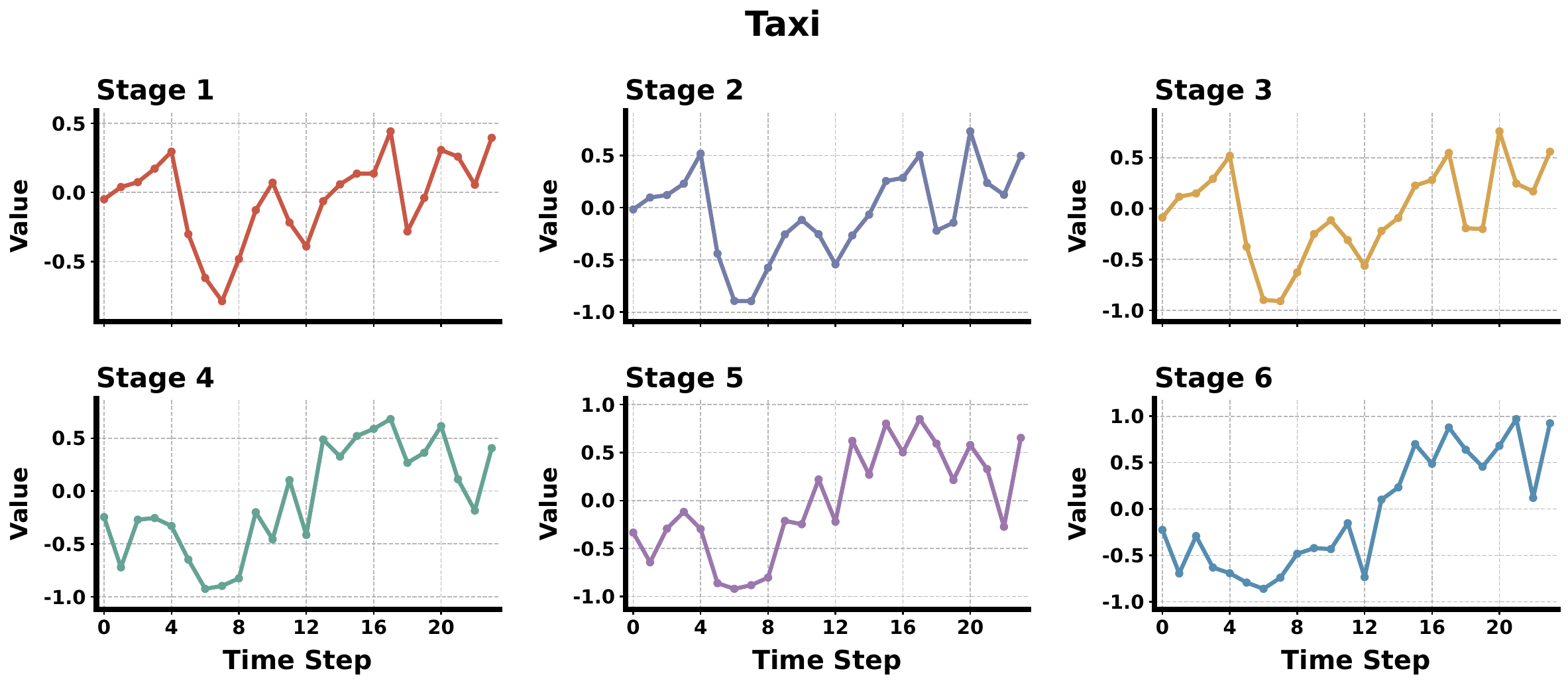}
    \caption{Autoregressive generation process guided by next frequency-band prediction on the Electricity and Taxi benchmarks with a sequence length of 24.}
    \label{fig:12}
\end{figure}

\noindent\textbf{Case Study: Next Frequency-Band Prediction.}
To provide a more intuitive understanding of the next frequency-band prediction process, we conduct a case study.
Figure~\ref{fig:12} illustrates the generation process on the Electricity and Taxi benchmarks. 
We observe that ORBIT first establishes a coarse-grained global temporal structure at the initial frequency level. 
As the generation progressively proceeds toward higher frequency bands, increasingly fine-grained local temporal details are incrementally incorporated into the generated sequence.

\end{document}